\documentclass[10pt,twocolumn,letterpaper]{article}

 \usepackage{cvpr}              

\usepackage[dvipsnames]{xcolor}

\definecolor{cvprblue}{rgb}{0.21,0.49,0.74}
\usepackage[pagebackref,breaklinks,colorlinks,citecolor=cvprblue]{hyperref}

\usepackage{algorithm}
\usepackage{algpseudocode}

\def\paperID{17062} 
\def\confName{CVPR}
\def\confYear{2024}

\title{Unsupervised Occupancy Learning from Sparse Point Cloud}

\author{%
  Amine Ouasfi \quad \quad Adnane Boukhayma\\
  Inria, Univ. Rennes, CNRS, IRISA, M2S, France
}

\begin{document}
\maketitle
\begin{abstract}
Implicit Neural Representations have gained prominence as a powerful framework for capturing complex data modalities, encompassing a wide range from 3D shapes to images and audio. Within the realm of 3D shape representation, Neural Signed Distance Functions (SDF) have demonstrated remarkable potential in faithfully encoding intricate shape geometry. However, learning SDFs from 3D point clouds in the absence of ground truth supervision remains a very challenging task. In this paper, we propose a method to infer occupancy fields instead of SDFs as they are easier to learn from sparse inputs. We leverage a margin-based uncertainty measure to differentiably sample from the decision boundary of the occupancy function and supervise the sampled boundary points using the input point cloud. We further stabilise the optimization process at the early stages of the training by biasing the occupancy function towards minimal entropy fields while maximizing its entropy at the input point cloud. Through extensive experiments and evaluations, we illustrate the efficacy of our proposed method, highlighting its capacity to improve implicit shape inference with respect to baselines and the state-of-the-art using synthetic and real data.
\end{abstract}    
\section{Introduction}
\label{sec:intro}

Capturing full 3D shape from limited and corrupted data is a long standing problem. In this regard, one of the data type instances that has received increasing attention and investigation from computer vision, graphics and machine learning is point clouds. This interest emanates primarily from the ubiquity of this light, although topologically incomplete 3D representation, either as acquired \eg through industrial and commodity depth sensors, or as an intermediate representation within computational photogrammetry \cite{schoenberger2016sfm,schoenberger2016mvs} pipelines for instance. 
While classical optimization methods such as Poisson Reconstruction \cite{kazhdan2013screened} or Moving Least Squares \cite{guennebaud2007algebraic} can be effective with dense, clean point sets and accurate normal pre-estimations,  recent deep learning-based alternatives provide more robust predictions, particularly for noisy and sparse inputs, eliminating the need for normal data in many cases. 

Several methods rely on priors learned from large fully labeled data such as the synthetic dataset ShapeNet \cite{shapenet}. However, this strategy entails computationally expensive trainings, and the resulting models can still be prone to out-of-distribution generalization issues, as pointed in \cite{NeuralTPS,ouasfi2023Robustifying}, whether caused by change in the input size or domain shift. For instance, Table
\ref{tab:fs} shows that our unsupervised method outperforms supervised generalizable models when testing on data that is sparser and different in nature from their training corpus. Hence, it is important to design learning frameworks that can lead to robust reconstruction under such extreme constraints.

Neural implicit representaions (INR) were established recently as a powerful representation for 3D shape, generally taking the form of coordinate based MLPs. They have been applied to model various shape presentations, the most popular ones of which remain arguably signed distance functions (SDF) \cite{park2019deepsdf} and binary occupancy fields \cite{mescheder2019occupancy} for watertight shapes.
For the task of point cloud reconstruction, there is a wide literature on learning SDFs from dense labels, as well as from point clouds solely (unsupervised), either in the generalizable setting or in instance specific optimization. However, apart from its applications in the supervised generalizable setting \cite{peng2020convolutional,boulch2022poco,chibane2020implicit}, it seems that occupancy learning from point sets has not been as widely considered in other scenarios. In particular, to the best of our knowledge, using occupancy fields to learn shapes from sparse noisy unoriented point clouds (\ie unsupervisedly) has not been explored thoroughly. This is even the more intriguing as it seems to be easier to learn a binary classification task as opposed to regressing a continuous field that must additionally respect specific properties \cite{gropp2020implicit}.  

Concordantly, we introduce occupancy in this work for learning shape from a sparse point cloud. While other representations such as SDF can readily incorporate surface sample supervision directly (\eg \cite{gropp2020implicit}), among other strategies, it is not entirely straightforward to define such supervision for occupancy fields. Hence, we propose a novel strategy for our training that combines a loss applied near the boundary decision, harnessing the input point samples, and an accompanying regularization. For the first loss, we encourage the decision boundary of our occupancy to align with the point cloud. We achieve this by supervising an uncertainty sampling mechanism for our occupancy field with the point cloud samples. We approximate this sampling through root-finding of a margin function associated to the field. As regularization, we propose to minimize the uncertainty of our field, while maximizing it near the decision boundary, via the entropy of our occupancy.

Through extensive experiments under several real and synthetic benchmarks for object, non-rigid and scene level shape reconstruction, our results show that our method out-performs the existing state-of-the-art literature in reconstruction from sparse point cloud, using standard metrics and through superior visual results.  
Our ablation studies validate our design choices and showcase the importance of the components of the supervision scheme that we propose.







\section{Related Work}
\label{sec:rel}

\noindent\textbf{Shape Representations in Deep Learning}
Shapes can be depicted in deep learning through either intrinsic or extrinsic representations. Intrinsic representations focus on discriminating the shape itself. When explicitly implemented, for instance, using tetrahedral or polygonal meshes \cite{wang2018pixel2mesh,kato2018neural,jena2022neural} or point clouds \cite{fan2017point,aliev2020neural}, the output topology is predefined, thereby restricting the range of shapes that can be generated. Among other intrinsic representations, 2D patches \cite{groueix2018papier,williams2019deep,deprelle2019learning} may introduce discontinuities, while the simplicity of shape primitives like cuboids \cite{abstractionTulsiani17,zou20173d}, planes \cite{liu2018planenet}, and Gaussians \cite{genova2019learning,kerbl20233d} constrains their expressiveness. On the other hand, extrinsic shape representations model the entire space containing the scene or object of interest. Voxel grids \cite{wu20153d,wu2016learning} are the most popular, serving as a direct extension of 2D pixels to the 3D domain. However, their capacity is constrained by the memory cost associated with cubic resolution. Sparse representations, such as octrees \cite{riegler2017octnet,tatarchenko2017octree,wang2017cnn}, can mitigate this issue to some extent.

\noindent\textbf{Implicit Neural Shape Representations}
Implicit Neural Shape Representations have recently emerged as a significant approach for modeling extrinsic shape, radiance and light fields (\eg \cite{mildenhall2020nerf,yariv2021volume,wang2021neus,jain2021dreamfields,chan2022efficient,li2023learning,li2023regularizing}). These representations address many of the limitations associated with classical representations by showcasing the ability to represent shapes with arbitrary topologies at virtually infinite resolution. Typically, they are parameterized with MLPs, which map spatial locations or features to properties such as occupancy \cite{mescheder2019occupancy}, signed \cite{park2019deepsdf}, or unsigned \cite{chibane2020neural,Zhou2022CAP-UDF} distances relative to the target shape. The level-set derived from these MLPs can be visualized through techniques like ray marching \cite{hart1996sphere} or tessellated into an explicit shape using methods like Marching Cubes \cite{lorensen1987marching}. Another notable line of research involves the development of hybrid implicit/explicit representations \cite{palmer2022deepcurrents,chen2020bsp,deng2020cvxnet,yavartanoo20213dias}, primarily based on differentiable space partitioning. To simultaneously represent collections of shapes, implicit neural models necessitate conditioning mechanisms. These mechanisms encompass features and latent code concatenation, batch normalization, hypernetworks \cite{sitzmann2020implicit,NEURIPS2019_b5dc4e5d,sitzmann2021light,wang2021metaavatar,chen2022transinr}, and gradient-based meta-learning \cite{ouasfi2022few,sitzmann2020metasdf}. Concatenation-based conditioning was initially implemented using single global latent codes \cite{mescheder2019occupancy,chen2019learning,park2019deepsdf} and has been further refined with the incorporation of local features \cite{li2022learning,genova2020local,tretschk2020patchnets,takikawa2021neural,peng2020convolutional,chibane2020implicit,jiang2020local,erler2020points2surf}.

\noindent\textbf{Reconstruction from Point Clouds}
Classical approaches to reconstruction include combinatorical methods where the shape is defined based on input point clouds through space partitioning, employing techniques such as alpha shapes \cite{bernardini1999ball}, Voronoi diagrams \cite{amenta2001power}, or triangulation \cite{cazals2006delaunay,liu2020meshing,rakotosaona2021differentiable}. Alternatively, the input samples can contribute to defining an implicit function, with its zero level set representing the target shape. This is achieved through global smoothing priors \cite{williams2022neural,lin2022surface,williams2021neural}, such as radial basis functions \cite{carr2001reconstruction} and Gaussian kernel fitting \cite{scholkopf2004kernel}, or local smoothing priors like moving least squares \cite{mercier2022moving,guennebaud2007algebraic,kolluri2008provably,liu2021deep}. Another approach involves solving a boundary-conditioned Poisson equation \cite{kazhdan2013screened}. Recent literature suggests parameterizing these implicit functions with deep neural networks and learning their parameters through gradient descent, either in a supervised or unsupervised manner.

\noindent\textbf{Supervised Implicit Neural Reconstruction}
In supervised methods, there is an assumption of having labeled training data, typically in the form of dense samples containing ground truth shape information. Auto-decoding techniques \cite{li2022learning,park2019deepsdf,tretschk2020patchnets,jiang2020local,chabra2020deep} necessitate test-time optimization to adapt to a new point cloud. On the other hand, encoder-decoder methods allow for swift feed-forward inference. Initially introduced for this purpose, Pooling-based set encoders \cite{mescheder2019occupancy,chen2019learning,genova2020local} like PointNet \cite{qi2017pointnet} have been found to underfit the context. State-of-the-art performance is achieved by convolutional encoders, utilizing local features defined either in explicit volumes and planes \cite{peng2020convolutional,chibane2020implicit,lionar2021dynamic,peng2021shape} or solely at the input points \cite{boulch2022poco,ouasfi2024Mixing}. Peng \etal. \cite{peng2021shape} proposed a differentiable Poisson solving layer that efficiently converts predicted normals into an indicator function grid. However, its applicability is limited to small scenes due to the cubic memory requirement in grid resolution. The work in \cite{williams2022neural,huang2023neural} suggests constructing generalizable models with implicit decoder functions as a kernel regression. Despite these advancements, many of the generalizable methods still face challenges related to generalization.

\noindent\textbf{Unsupervised Implicit Neural Reconstruction}
For unsupervised approaches, a neural network is typically fitted to the input point cloud without additional information. Convergence improvements can be achieved through regularization techniques, such as the spatial gradient constraint based on the Eikonal equation introduced by Gropp et al. \cite{gropp2020implicit}, a spatial divergence constraint as described in \cite{ben2022digs}, and Lipschitz regularization on the network \cite{liu2022learning}. Periodic activations were introduced in \cite{sitzmann2020implicit}. Lipman \cite{lipman2021phase} learns a function that converges to occupancy, while its log transform converges to a distance function. Atzmon \etal \cite{atzmon2020sal} learn an SDF from unsigned distances, further supervising the spatial gradient of the function with normals \cite{atzmon2020sald}. Ma \etal \cite{ma2020neural} express the nearest point on the surface as a function of the neural signed distance and its gradient. They also utilize self-supervised local priors to handle very sparse inputs \cite{ma2022reconstructing} and enhance generalization \cite{ma2022surface}. \cite{koneputugodage2023octree} guides the implicit field learning with an Octree based labelling. \cite{boulch2021needrop} predicts occupancy fields by learning whether a dropped needle goes across the surface or no. \cite{NeuralTPS} learns a surface parameterization leveraged to provide additional coarse surface supervision to the shape network. In \cite{williams2021neural}, infinitely wide shallow MLPs are learned as random feature kernels using points and their normals. However, most of the methods mentioned above encounter challenges when dealing with sparse and noisy input, primarily due to the lack of supervision. Differently from this literature, we propose here to apply occupancy to unsupervised sparse point cloud neural fitting for the first time to the best of our knowledge.

\section{Method}
\label{sec:meth}

Given a sparse noisy unoriented input point cloud $\mathcal{P}\subset \mathbb{R}^{3\times N}$, our goal is to recover a  3D watertight shape surface $\mathcal{S}$ that best explains this observation, \ie points from  $\mathcal{P}$ approximating noisy samples from $\mathcal{S}$. 

Based on the proven success of implicit neural representations, we propose to address this problem by fitting an implicit neural shape function to the scarce observation, which amounts to learning an MLP mapping coordinates to shape attributes. In this context, dubbed unsupervised, signed distance fields  are the representation of choice in the community so far (\eg \cite{NeuralTPS,ma2020neural,chen2023gridpull,koneputugodage2023octree,peng2021shape,williams2021neural}). Oddly enough, binary occupancy fields were not explored extensively yet as a representation for this context to the best of our knowledge, even-though in theory,  it seems \textit{easier} to learn occupancy than SDFs. In fact, it is safe to assume that binary classification is an easier task than continuous regression in general. Besides, we can see intuitively that occupancy is only the \textit{sign} part of the SDF information. Additionally, occupancy fields do not seem to stipulate any strong structural requirement, as it is the case for distance fields, whose gradient ought to satisfy the Eikonal constraint \cite{gropp2020implicit} for instance. Hence, we propose here to learn a binary occupancy field.

When labeled samples inside and outside the shape are available, such a neural occupancy field could be learned in this supervised setting via a standard cross-entropy classification loss (\eg \cite{mescheder2019occupancy}). However when only surface points are available, as in our case ($\mathcal{P}$), it is not entirely clear how a successful learning can be achieved using such occupancy decision boundary samples alone. We derive in the following a methodology to approach this task. 

\subsection{Learning Occupancy Through Margin Uncertainty Sampling}
\label{sec:samp}

We are interested in learning the following posterior $P(y|\bold{x},\theta)$, where $\bold{x}$ represents query points in $\mathbb{R}^3$, and label $y\in\{0,1\}$ stands for binary shape occupancy. $P$ is parameterized with a Softmax activated MLP $\theta$ similarly to \cite{boulch2022poco}. Since we do not have points with their occupancy labels, all we can learn with is the input point cloud samples. 

Let us consider the following margin function:
\begin{equation}
\label{equ:U_t}
    U_{\theta}(\bold{x}) = P(y=1|\bold{x},\theta)- P(y=0|\bold{x},\theta).
\end{equation}
Borrowing from active learning terminology, finding the roots of  this expression is akin to a form of uncertainty sampling from our occupancy field (margin sampling \cite{settles2009active}). Intuitively, the smaller the margin of a point the more uncertain its shape prediction. In this particular case, the margin function zeroes correspond to the samples with upmost level of uncertainty.

Given that our input point cloud points are samples from the surface, \ie the ground truth occupancy function decision boundary, we propose to learn our occupancy field by supervising uncertainty sampling with the point cloud, \ie minimizing the distance between uncertain samples and their nearest point cloud samples. 
We define our uncertainty sampling as root finding of our margin function. We Initialize this root finding near the target surface. We hypothesis that training our occupancy field by making its uncertain samples coincide with the surface through minimal amount of steps of a root finding algorithm will encourage it to converge quickly and decisively.  

We start by generating a pool of query points near $\mathcal{P}$, \ie near the ground-truth surface, by sampling around the points, \ie $\{\bold{q} \sim \mathcal{N}(\bold{p},\sigma_{\bold{p}} \mathbf{I}_3)\}$. Here, $\sigma_{\bold{p}}$ is chosen as the maximal euclidean distance to the $K$ nearest points to $\bold{q}$ in $\mathcal{P}$ (as in \cite{gropp2020implicit}). We subsequently recompute the nearest point $\bold{p}$ in $\mathcal{P}$ to each sample $\bold{q}$, thus forming the following set of training pairs:
\begin{equation}
\mathcal{Q} := \{(\bold{q},\bold{p}), \bold{p}=
\underset{\bold{v}\in\mathcal{P}}{\mathrm{argmin}}||\bold{v}-\bold{q}||_2\}.
\end{equation}

Given a pair $(\bold{q},\bold{p})$ in $\mathcal{Q}$, we train by bounding one step of Newton-Raphson \cite{ypma1995historical} root finding on our occupancy margin, initialized at $\bold{q}$, to result in an uncertain sample near $\bold{p}$ in mean squared error terms.

Let us recall that a Generalized Newton \cite{ben1966newton} (Theorem 1, with $m=1$ and $n=3$) iteration updates query point $\bold{q}$ accordingly: $\bold{q} - \nabla U_\theta(\bold{q})^\dagger U_\theta(\bold{q})$, where $\nabla U_\theta(\bold{q})^\dagger$ is the Moore-Penrose pseudoinverse of Jacobian $ \nabla U_\theta(\bold{q})$.  As $U_\theta$ is a scalar function ($\mathbb{R}^3\rightarrow \mathbb{R}$), $\nabla U_\theta$ is a row 3-vector (we can use it or its transpose interchangeably), and the pseudoinverse writes: $\nabla U_\theta(\bold{q})^\dagger = \frac{\nabla U_\theta(\bold{q})}{||\nabla U_\theta(\bold{q})||_2^2}$. 

In conclusion, our margin uncertainty sampling loss can be expressed as follows: 
\begin{equation}
\mathcal{L} _{\text{samp}} (\theta , \mathcal{Q} ) = 
\underset{\begin{subarray}{c}
(\bold{q},\bold{p}) \sim \mathcal{Q}\\
\end{subarray}}{\mathbb{E}}||\bold{q}- U_\theta(\bold{q}) \cdot \frac{\nabla U_\theta(\bold{q})}{||\nabla U_\theta(\bold{q})||_2^2} - \bold{p}||_2^2,   
\label{equ:samp}
\end{equation}
where spatial gradient $\nabla U_\theta$ can be computed efficiently through automatic differentiation (\eg PyTorch \cite{paszke2019pytorch}). 

This loss offers two primary advantages. Firstly, it serves as a form of supervision for the occupancy of the query points. The sign of the margin function differs on opposite sides of the decision boundary. Consequently, based on this sign, a Newton-Raphson iteration will move points along or against the direction of its gradient resulting in an occupancy field whose decision boundary is aligned with the input point cloud.  Additionally, this loss operates as a smoothness constraint as  Newton-Raphson root finding relies on approximating the function at the roots by its first-order approximation on the initial points $\bold{q}$, hence smoothing the margin function around the input points.

\subsection{Entropy Based Regularization}
\label{sec:entr}

To ensure better convergence of our training, we use the MLP initialization in \cite{atzmon2020sal}, while adapting it such that the last layer predicts binary occupancy for an $r$-radius sphere. 

Furthermore, to ease and improve our learning, we suggest to urge our occupancy field to have minimal uncertainty almost everywhere in space. We use the entropy \cite{shannon1948mathematical} as an uncertainty measure here. 
However, this is the opposite behavior that we expect from our network to display at the decision boundary \ie near the ground-truth surface. Hence, we propose to maximize the uncertainty (\ie entropy) conversely at the input point cloud $\mathcal{P}$ simultaneously. 

Given a random variable $X$, Shannon entropy \cite{shannon1948mathematical} is defined as follows:
\begin{equation}
\mathbb{H} (X) = - \mathbb{E} \log(P (X) ). 
\end{equation}

We create a set $\mathit{\Omega} \subset \mathbb{R}^3$ of query points $\bold{x}$ by sampling space uniformly within a bounding box of input point cloud $\mathcal{P}$. As a measure of uncertainty we consider the entropy of the occupancy distribution conditioned on the observed spatial location:
\begin{equation}
\mathcal{L}_{\text{entr}}(\theta,\mathit{\Omega},\mathcal{P}) = \mathbb{E}_{\bold{x} \sim \mathit{\Omega}}  \mathbb{H} (y |\bold{x},\theta ) - \mathbb{E}_{\bold{p} \sim \mathcal{P}}  \mathbb{H} (y |\bold{p},\theta ).
\label{equ:entr}
\end{equation}


This entropy polarization loss acts both as an initialization of the field and a regularization mid-learning, as we empirically found that it performs best when its weight ($ \lambda $ in  Equation \ref{eq:erm}) is reduced progressively in the final loss in the midst of training. 

\begin{figure}[t!]
\centering
\includegraphics[width=0.9\linewidth]{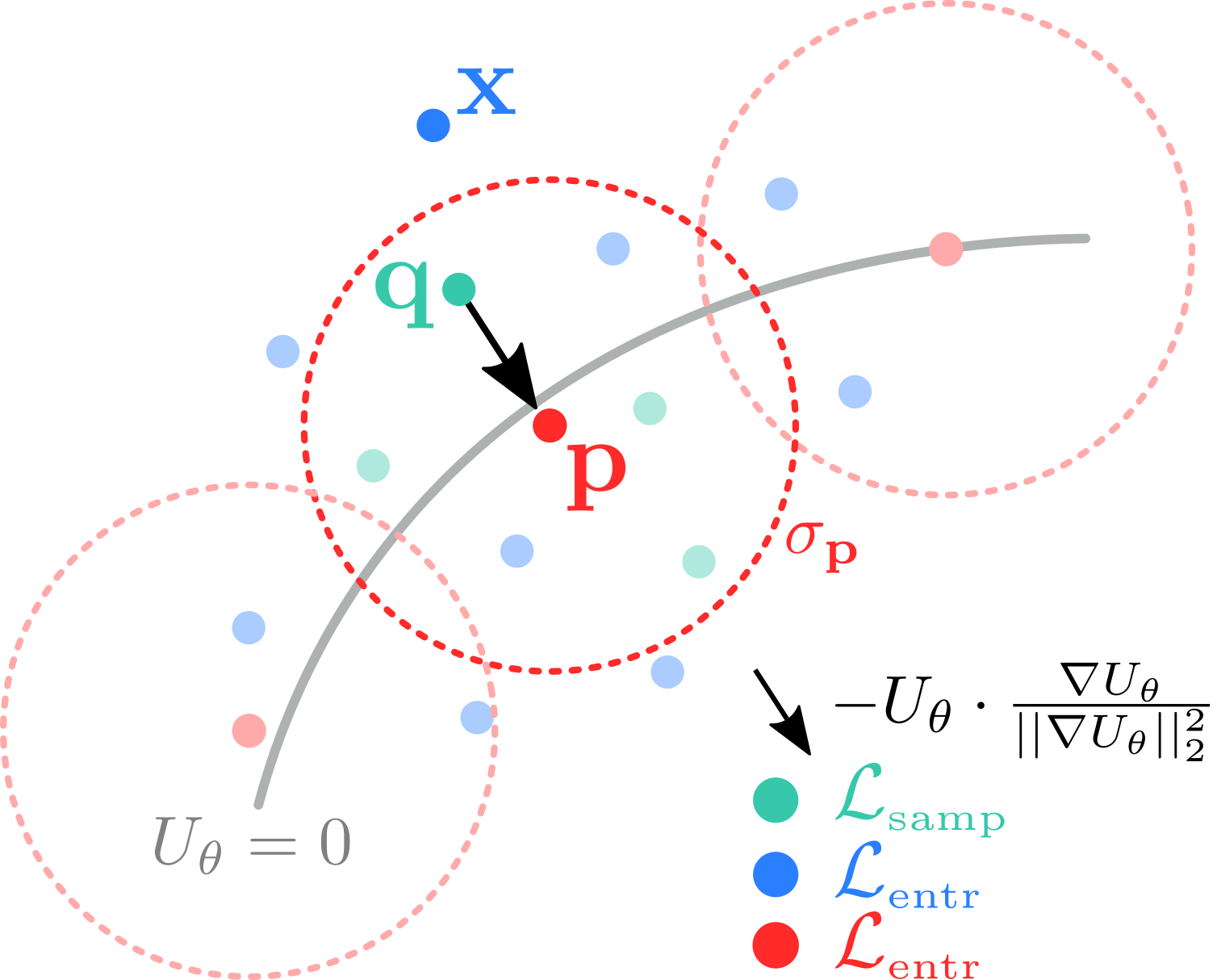}
\caption{Illustration of our training. Our method learns a neural binary occupancy field without off-surface labels. It uses the combination of a margin uncertainty sampling loss near the surface (Green), maximizing entropy at the input point cloud samples (Red), and minimizing entropy everywhere else (Blue).}
\label{fig:loss}
\end{figure}

Finally we can train our model using the following combined empirical risk minimization: 
\begin{equation}
\min_{\theta} 
\mathcal{L} _{\text{samp}} (\theta , \mathcal{Q} ) + 
 \lambda \mathcal{L}_{\text{entr}}(\theta,\mathit{\Omega},\mathcal{P}).
 \label{eq:erm}
\end{equation}

Algorithm \ref{alg:train} shows a summary of our training procedure. We note that once an implicit occupancy field is learned, an explicit triangle mesh $\hat{\mathcal{S}}$ can be obtained with it through the Marching Cubes algorithm \cite{lorensen1987marching}.

\begin{algorithm}
\small
\begin{algorithmic}
\Require Point cloud $\mathcal{P}$, learning rate $\alpha$, number of iterations $N_{\text{it}}$, number of near surface queries $N_\mathcal{Q}$, number of uniform queries $N_\mathit{\Omega}$, number of batch input points $N_{\mathcal{P}}$, number of uniform queries $N_{\mathit{\Omega}}$, reg. loss weight $\lambda$. 
\Ensure Optimal weights ${\theta}^*$.
\State Compute local st. dev. $\{\sigma_\bold{p}=\max_{\bold{v}\in K\text{nn}(\bold{p},\mathbf{P})}||\bold{v}-\bold{p}||_2\}$.
\State Generate sets $\mathcal{Q}$ (Sec.\ref{sec:samp}) and 
$\mathit{\Omega}$ (Sec.\ref{sec:entr}). 
\State Initialize $\lambda$ (Sec.\ref{sec:imp}).
\For {$N_{\text{it}}$ times}
\State Sample a batch $\mathcal{Q}_b$ of size $N_{\mathcal{Q}}$  from $\mathcal{Q}$.
\State Sample a batch $\mathit{\Omega}_b$ of size $N_{\mathit{\Omega}}$  from $,\mathit{\Omega}_b$.
 \State Sample a batch $\mathcal{P}_b$ of size $N_{\mathcal{P}}$  from $\mathcal{P}$.
\State Compute losses $\mathcal{L} _{\text{samp}} (\theta , \mathcal{Q}_b ) $ (Equ.\ref{equ:samp}).
\State Compute losses $\mathcal{L}_{\text{entr}}(\theta,\mathit{\Omega}_b,\mathcal{P}_b)$ (Equ.\ref{equ:entr}).
\State $\theta \leftarrow \theta - \alpha\nabla_\theta\left(
\mathcal{L}_{\text{samp}} + \lambda  \mathcal{L}_{\text{entr}} \right)$.
\State Update $\lambda$ (Sec.\ref{sec:imp}).
\EndFor
\end{algorithmic}
\caption{\small The training procedure of our method.}
\label{alg:train}
\end{algorithm}

\section{Implementation Details}
\label{sec:imp}

We set number of queries as $N_\mathcal{Q} = 1000k$ and $N_\mathit{\Omega} = 10k$. 
Our MLP follows the architecture in \cite{ma2020neural}, and similarly to the latter we set $K=51$. We train for $N_{it}=40k$ iterations on a Nvidia RTX A6000  GPU using the Adam optimizer with learning rate $\alpha=0.001 $. Our training takes roughly $5$ minutes for a $1024$ sized input point cloud.
The regularization parameter $\lambda$ (Equation \ref{eq:erm}) follows an exponential decay of the form $\exp{-\kappa t}$ where $\kappa$ is set to $1.84\times10^{-2}$.

\section{Results}
\label{sec:res}
To test the efficacy of our approach, we assess our capacity to infer implicit representations of shapes when presented with sparse and noisy point clouds. We employ datasets  from standard reconstruction benchmarks, showcasing a diverse array of challenges associated with implicit shape function inference from  sparse inputs. Consistently with existing literature, we evaluate the performance of our method by quantifying the accuracy of 3D explicit shape models obtained post-convergence from our MLPs. It is worth noting that our approach entails fitting an MLP independently for each point cloud, operating without reliance on pre-learned priors or additional training data. We  compare  quantitatively  and  qualitatively to the the state-of-the-art in our problem setting, \ie unsupervised reconstruction from unoriented point, including deep learning methods N-Pull \cite{ma2020neural}, NTPS \cite{NeuralTPS}, G-Pull \cite{chen2023gridpull}, OG-INR \cite{koneputugodage2023octree}, DiGs \cite{ben2022digs},  NDrop \cite{boulch2021needrop}, SAP \cite{peng2021shape}. We show results for NSpline \cite{williams2021neural} even-though it requires normals. We also compare to classical Poisson Reconstruction (SPSR \cite{kazhdan2013screened}). We note that NTPS is the closest method to ours as it focuses specifically on the sparse input case. 
For comprehensive evaluation, we also include comparisons with feed-forward generalizable methods, namely POCO \cite{boulch2022poco} and CONet \cite{peng2020convolutional}, alongside the prior-based optimization method On-Surf  \cite{ma2022reconstructing} dedicated to sparse inputs. Unless stated differently, we use the publicly available official implementations of existing methods.

\begin{table}[t!]
\centering
\begin{tabular}{lllll}
\hline
& CD1     & CD2    & NC       & FS  \\ \hline
SPSR  \cite{kazhdan2013screened}   & 2.34          & 0.224          & 0.74          & 0.50         \\
OG-INR \cite{koneputugodage2023octree} & 1.36          & 0.051          & 0.55          & 0.55          \\
N-Pull \cite{ma2020neural}    & 1.16          & 0.074          & 0.84          & 0.75          \\
G-Pull \cite{chen2023gridpull} &1.07	&0.032	& 0.70 & 0.74\\
NTPS \cite{NeuralTPS}   & 1.11          & 0.067          & \textbf{0.88} & 0.74         \\
Ours   & \textbf{0.76} & \textbf{0.020} & \textbf{0.88}          & \textbf{0.83} \\ \hline
\end{tabular}
\caption{ShapeNet \cite{shapenet} reconstructions from sparse noisy unoriented point clouds.}
\label{tab:sn}
\end{table}

\begin{figure*}[t!]
\centering
\includegraphics[width=.7\linewidth]{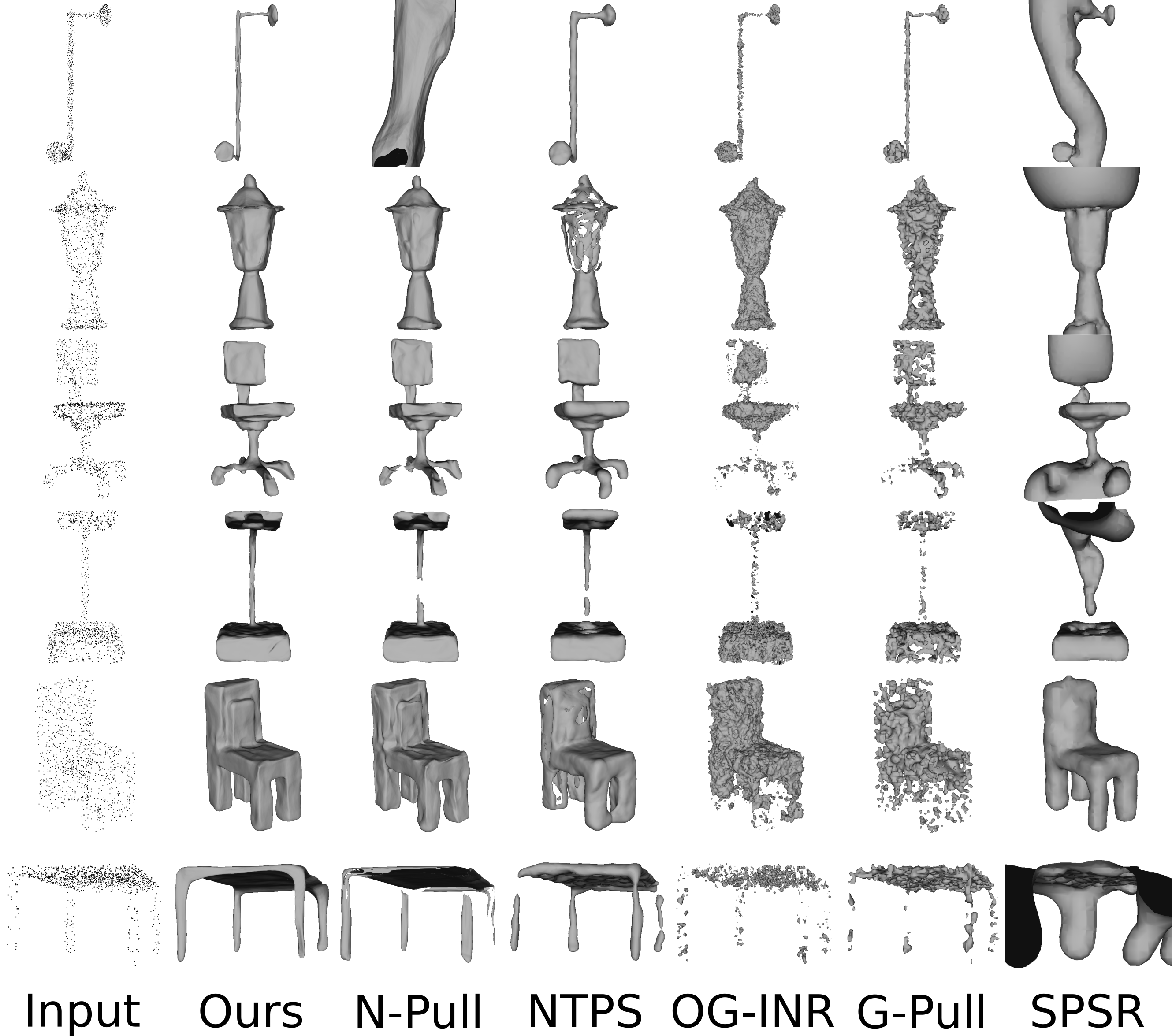}
\caption{ShapeNet \cite{shapenet} reconstructions from sparse noisy unoriented point clouds.}
\label{fig:sn}
\end{figure*}

\subsection{Metrics}

Following seminal work, we evaluate our method and the competition \wrt the ground truth using standard metrics for the 3D reconstruction task. Namely, the L1 \textbf{Chamfer Distance
(CD1)} ($\times10^{2}$), L2 \textbf{Chamfer Distance
(CD2)} ($\times10^{2}$), the \textbf{Hausdorff distance (HD)} and the euclidean distance based \textbf{F-Score (FS)} when ground truth points are available, and finally \textbf{Normal Consistency (NC)} when ground truth normals are available. We detail the expressions of these metrics in the supplementary material.

\subsection{Datasets and Input Definitions}

\textbf{ShapeNet} \cite{shapenet} consists of various instances of 13 different synthetic 3D object classes. We follow the train/test splits defined in \cite{williams2021neural}. We generate noisy input point clouds by sampling $1024$ points from the meshes and adding Gaussian noise of standard deviation $0.005$ following the literature (\eg \cite{boulch2022poco,peng2020convolutional}). For brevity we show results on classes Tables, Chairs and Lamps. \textbf{Faust} \cite{Bogo:CVPR:2014} consists of real scans of 10 human body identities in 10 different poses. We sample sets of $1024$ points from the scans as inputs. \textbf{3D Scene} \cite{zhou2013dense} contains large scale complex real world scenes obtained with a handheld commodity range sensor. We follow \cite{NeuralTPS,jiang2020local,ma2020neural} and sample our input point clouds with a sparse density of $100$ per m$^2$, and we report performance similarly for scenes Burghers, Copyroom, Lounge, Stonewall and Totempole. \textbf{Surface Reconstruction Benchmark (SRB)} \cite{williams2019deep} consists of five object scans, each with different challenges such as complex topology, high level of detail, missing data and varying feature scales. We sample $1024$ points from the scans for the sparse input experiment, and we also experiment using the dense inputs.

\subsection{Object Level Reconstruction}

We conduct the reconstruction of ShapeNet \cite{shapenet} objects using sparse and noisy point clouds. Table \ref{tab:sn} and Figure \ref{fig:sn} present a quantitative and qualitative comparison with other methods. Our approach consistently outperforms the competition across all metrics, as seen in the visual superiority of our reconstructions. We excel in capturing fine structures and details with greater fidelity. While NTPS demonstrates good overall coarse reconstructions, its thin plate spline smoothing prior appears to limit its expressivity. Additionally, OG-INR struggles to achieve satisfactory results in the sparse and noisy regime, despite its effective Octree-based sign field guidance in denser settings.

\begin{table}[t!]
\centering
\begin{tabular}{lllll}
\hline
& CD1     & CD2    & NC       & FS  \\ \hline
POCO \cite{boulch2022poco} & 0.308     & 0.002   & 0.934        & 0.981    \\
CONet  \cite{peng2020convolutional}  & 1.26     & 0.028    & 0.829      & 0.599    \\
On-Surf \cite{ma2022reconstructing} &0.584    &0.012   &0.936 & 0.915  \\
\hline
SPSR  \cite{kazhdan2013screened}  & 0.751     & 0.028    & 0.871        & 0.839    \\
G-Pull \cite{chen2023gridpull} &0.495    & 0.005   &0.887 & 0.945\\
NTPS \cite{NeuralTPS}   & 0.737     & 0.015     & 0.943          & 0.844     \\
Ours  & \textbf{0.260}    & \textbf{0.002}     & \textbf{0.952} & \textbf{0.974}    \\ \hline
\end{tabular}
\caption{Faust \cite{Bogo:CVPR:2014} reconstructions from sparse noisy unoriented point clouds.}
\label{tab:fs}
\end{table}

\begin{figure}[t!]
\centering
\includegraphics[width=1.0\linewidth]{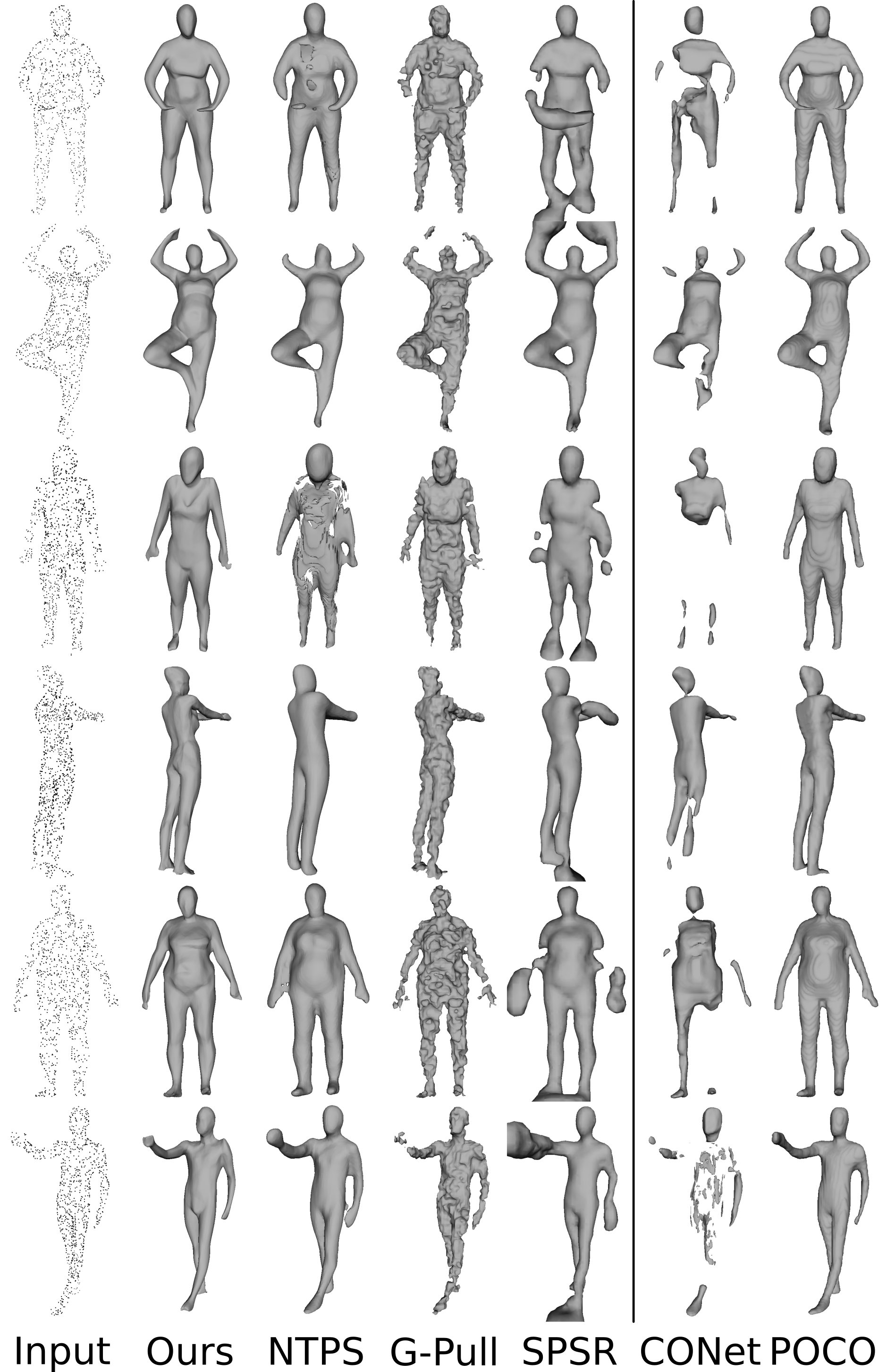}
\caption{Faust \cite{Bogo:CVPR:2014} reconstructions from sparse unoriented point clouds.}
\label{fig:fs}
\end{figure}

\subsection{Real Articulated Shape Reconstruction}
We undertake the reconstruction of Faust \cite{Bogo:CVPR:2014} human shapes using sparse and noisy point clouds. Table \ref{tab:fs} and Figure \ref{fig:fs} provide a numerical and qualitative comparison with other methods. Our approach surpasses the alternatives across all metrics, with visually superior reconstructions, as especially noticeable at the extremities of the body. Analogous to the fine structures in the ShapeNet experiment, these areas pose challenges due to limited input point cloud samples, making shape prediction inherently difficult and ambiguous. NTPS reconstructions, similarly, exhibit coarser and less detailed results on this dataset.  It is worth highlighting that our method outperforms not only feed-forward generalizable approaches POCO and  CONet but also the prior-based optimization method On-Surf. These methods rely on priors trained on ShapeNet, thereby constraining their generalization ability.

\begin{table*}[h!]
\centering
\scalebox{0.75}{
\begin{tabular}{l|lllllllllllllllllll}
\cline{1-19}
\multicolumn{1}{c|}{} & \multicolumn{3}{c|}{Burghers}                                                       & \multicolumn{3}{c|}{Copyroom}                                                       & \multicolumn{3}{c|}{Lounge}                                                         & \multicolumn{3}{c|}{Stonewall}                                                      & \multicolumn{3}{c|}{Totemple}                                                       & \multicolumn{3}{c}{\textbf{Mean}}                                    &  \\ \cline{2-19}
\multicolumn{1}{c|}{}                  & \multicolumn{1}{c}{CD1}  & \multicolumn{1}{c}{CD2}  & \multicolumn{1}{c|}{NC}    & \multicolumn{1}{c}{CD1}  & \multicolumn{1}{c}{CD2}  & \multicolumn{1}{c|}{NC}    & \multicolumn{1}{c}{CD1}  & \multicolumn{1}{c}{CD2}  & \multicolumn{1}{c|}{NC}    & \multicolumn{1}{c}{CD1}  & \multicolumn{1}{c}{CD2}  & \multicolumn{1}{c|}{NC}    & \multicolumn{1}{c}{CD1}  & \multicolumn{1}{c}{CD2}  & \multicolumn{1}{c|}{NC}    & \multicolumn{1}{c}{CD1} & \multicolumn{1}{c}{CD2} & \multicolumn{1}{c}{NC}\\ \cline{1-19}
SPSR   \cite{kazhdan2013screened}                                & 0.178                     & 0.205                     & \multicolumn{1}{l|}{0.874} & 0.225                     & 0.286                     & \multicolumn{1}{l|}{0.861} & 0.280                     & 0.365                     & \multicolumn{1}{l|}{0.869} & 0.300                     & 0.480                     & \multicolumn{1}{l|}{0.866} & 0.588                     & 1.673                     & \multicolumn{1}{l|}{0.879} & 0.314                    & 0.602                    & 0.870 &  \\
NDrop  \cite{boulch2021needrop}                                & 0.200                     & 0.114                     & \multicolumn{1}{l|}{0.825} & 0.168                     & 0.063                     & \multicolumn{1}{l|}{0.696} & 0.156                     & 0.050                     & \multicolumn{1}{l|}{0.663} & 0.150                     & 0.081                     & \multicolumn{1}{l|}{0.815} & 0.203                     & 0.139                     & \multicolumn{1}{l|}{0.844} & 0.175                    & 0.089                    & 0.769 &  \\
N-Pull   \cite{ma2020neural}                                  & 0.064                     & 0.008                     & \multicolumn{1}{l|}{0.898} & 0.049                     & 0.005                     & \multicolumn{1}{l|}{0.828} & 0.133                     & 0.038                     & \multicolumn{1}{l|}{0.847} & 0.060                     & 0.005                     & \multicolumn{1}{l|}{0.910} & 0.178                     & 0.024                     & \multicolumn{1}{l|}{0.908} & 0.097                    & 0.016                    & 0.878 &  \\
SAP  \cite{peng2021shape}                                  & 0.153                     & 0.101                     & \multicolumn{1}{l|}{0.807} & 0.053                     & 0.009                     & \multicolumn{1}{l|}{0.771} & 0.134                     & 0.033                     & \multicolumn{1}{l|}{0.813} & 0.070                     & 0.007                     & \multicolumn{1}{l|}{0.867} & 0.474                     & 0.382                     & \multicolumn{1}{l|}{0.725} & 0.151                    & 0.100                    & 0.797 &  \\
NSpline  \cite{williams2021neural}                              & 0.135                     & 0.123                     & \multicolumn{1}{l|}{0.891} & 0.056                     & 0.023                     & \multicolumn{1}{l|}{0.855} & 0.063                     & 0.039                     & \multicolumn{1}{l|}{0.827} & 0.124                     & 0.091                     & \multicolumn{1}{l|}{0.897} & 0.378                     & 0.768                     & \multicolumn{1}{l|}{0.892} & 0.151                    & 0.209                    & 0.88  &  \\
NTPS    \cite{NeuralTPS}                               & 0.055                     & 0.005                     & \multicolumn{1}{l|}{\textbf{0.909}} & 0.045                     & 0.003                     & \multicolumn{1}{l|}{\textbf{0.892}} & 0.129                     & 0.022                     & \multicolumn{1}{l|}{\textbf{0.872}} & 0.054                     & 0.004                     & \multicolumn{1}{l|}{\textbf{0.939}} & 0.103                     & 0.017                     & \multicolumn{1}{l|}{0.935} & 0.077                    & 0.010                    & \textbf{0.897} &  \\ \cline{1-19}
Ours                                   & \multicolumn{1}{r}{\textbf{0.022}} & \multicolumn{1}{r}{\textbf{0.001}} & \multicolumn{1}{l|}{0.871}  & \multicolumn{1}{r}{\textbf{0.041}} & \multicolumn{1}{r}{\textbf{0.012}} & \multicolumn{1}{l|}{0.812}  & \multicolumn{1}{r}{\textbf{0.021}} & \multicolumn{1}{r}{\textbf{0.001}} & \multicolumn{1}{l|}{0.870}  & \multicolumn{1}{r}{\textbf{0.028}} & \multicolumn{1}{r}{\textbf{0.003}} & \multicolumn{1}{l|}{0.931}  & \multicolumn{1}{r}{\textbf{0.026}} & \multicolumn{1}{r}{\textbf{0.001}} & \multicolumn{1}{l|}{\textbf{0.936}} & \textbf{0.027}                    & \textbf{0.003}                    & 0.886 &  \\ \cline{1-19}
\end{tabular}}
\caption{3D Scene \cite{zhou2013dense} reconstructions from sparse  point clouds.}
\label{tab:3ds}
\end{table*}

\begin{figure}[t!]
\centering
\includegraphics[width=.6\linewidth]{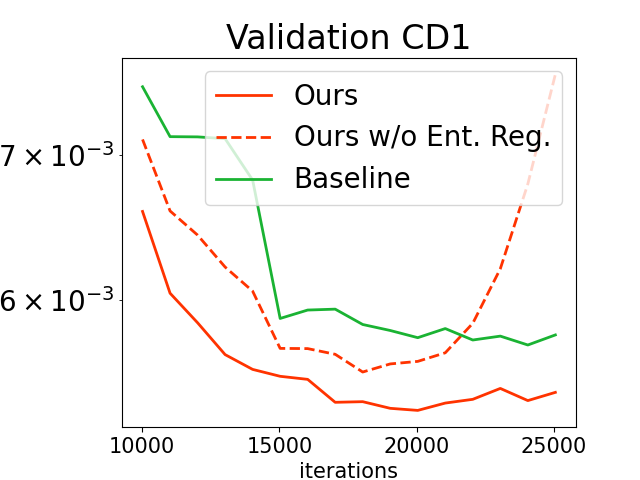}
\vspace{-10pt}
\caption{CD1 distance to GT for reconstructions of shape Gargoyle of benchmark SRB \cite{williams2019deep} from a sparse unoriented point cloud.}
\label{fig:loss}
\end{figure}

\begin{figure*}[t!]
\centering
\includegraphics[width=0.7\linewidth]{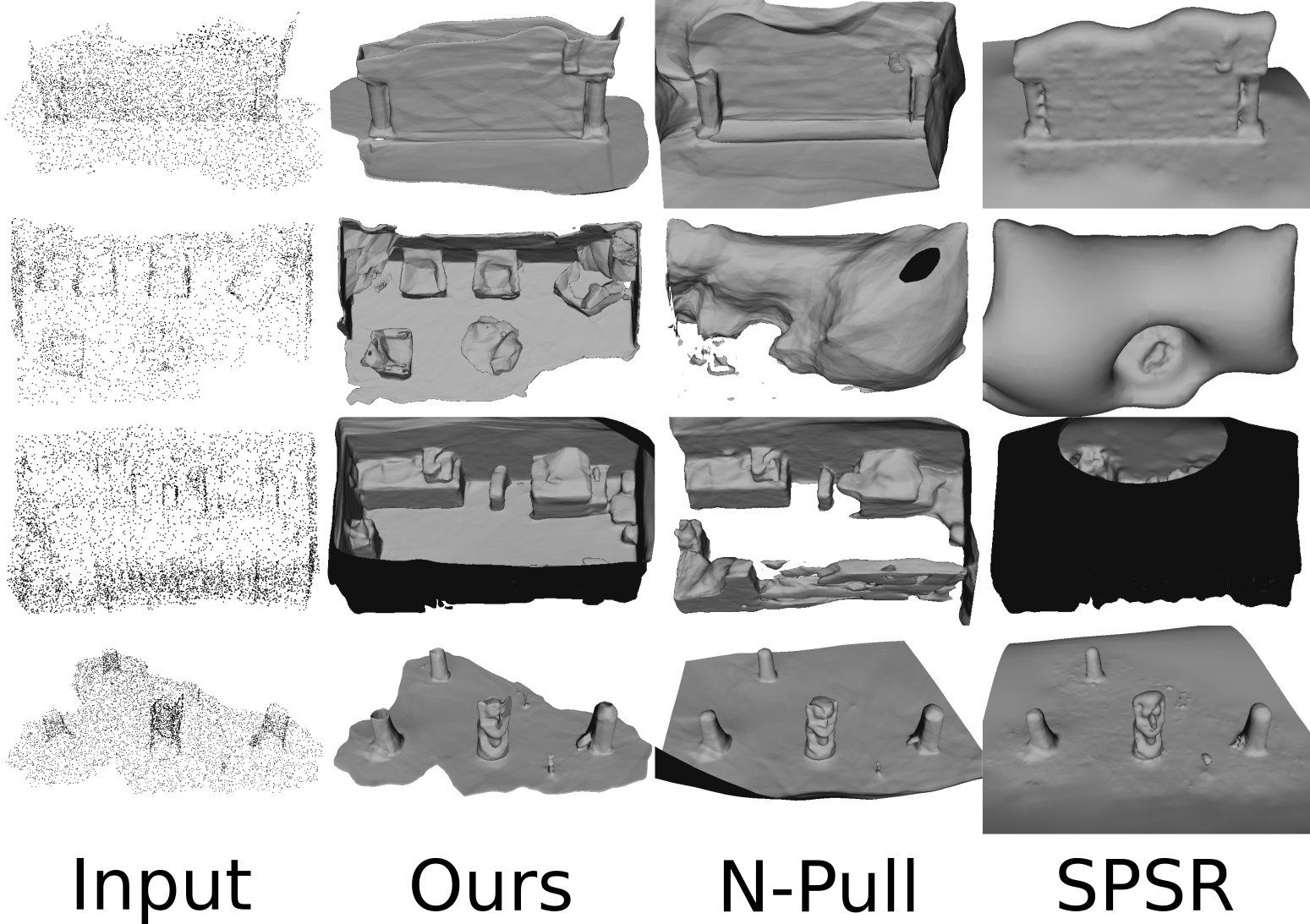}
\caption{3D Scene \cite{zhou2013dense} reconstructions from sparse unoriented point clouds.}
\label{fig:3ds}
\end{figure*}

\subsection{Real Scene Level Reconstruction}
In accordance with \cite{NeuralTPS}, we present the reconstruction outcomes on the 3D Scene \cite{zhou2013dense} dataset derived from spatially sparse point clouds. Table \ref{tab:3ds} provides a summary of numerical results, incorporating results for methods NTPS, N-Pull, SAP, NDrop, and NSpline as reported in the state-of-the-art method NTPS. Our method excels in this benchmark, surpassing the competition in most metrics. Notably, our baseline N-Pull exhibits more conspicuous failures in this extensive sparse setup. Figure \ref{fig:3ds} offers qualitative comparisons against baselines N-Pull and SPSR.

\section{Ablation Studies}
\label{sec:abl}


\begin{table}[h]
\centering
\scalebox{0.8}{
\begin{tabular}{lll}
\hline
\textbf{}                & CD1             & NC   \\ \hline
N-Pull \cite{ma2020neural} (baseline)                   & 1.10          & 0.85 \\ 
$\mathcal{L}_{\text{samp}}$ (Equ.\ref{equ:samp})   & 1.02         & 0.86 \\
$\mathcal{L}_{\text{samp}}$ + Margin reg   & 0.80 & 0.88 \\
$\mathcal{L}_{\text{samp}}$ +  SA-Occ \cite{tang2021sa} reg   & 0.95          & 0.88 \\
Ours          & \textbf{0.77}  & \textbf{0.89} \\ \hline
\end{tabular}}
\caption{Ablation of our method on the Tables class of  ShapeNet  \cite{shapenet} .}
\label{tab:abl}
\end{table}

\textbf{Loss}. Our approach centers around aligning the decision boundary of our occupancy function with the shape surface through a loss function based on uncertainty sampling, denoted as $\mathcal{L}_{\text{samp}}$ (Equation \ref{equ:samp}). To facilitate and enhance our learning process, we impose constraints on the network to minimize uncertainty almost everywhere in space while maximizing it at the decision boundary. Entropy serves as our chosen uncertainty measure. However, other measures can be considered. We justify this choice in Table \ref{tab:abl}. We compare our entropy-based regularization to margin-based regularization (Margin reg), using the absolute value of our margin function $U_{\theta}$ (Equation \ref{equ:samp}) as an uncertainty measure instead of entropy. Additionally, we compare it to the sign-agnostic occupancy supervision strategy introduced in \cite{tang2021sa} denoted as SA-Occ. Although this strategy fails to converge in the absence of priors, we demonstrate its beneficial use as a regularization in place of our entropy-based approach. It is noteworthy that while margin-based regularization (Margin reg) yields better results, it is still outperformed by our entropy-based regularization. Figure \ref{fig:loss} shows an example of validation plots, displaying the role of our regularization in avoiding overfitting. The baseline in this example is SDF method N-Pull \cite{ma2020neural}.

\begin{figure}[t!]
\centering
\includegraphics[width=1.0\linewidth]{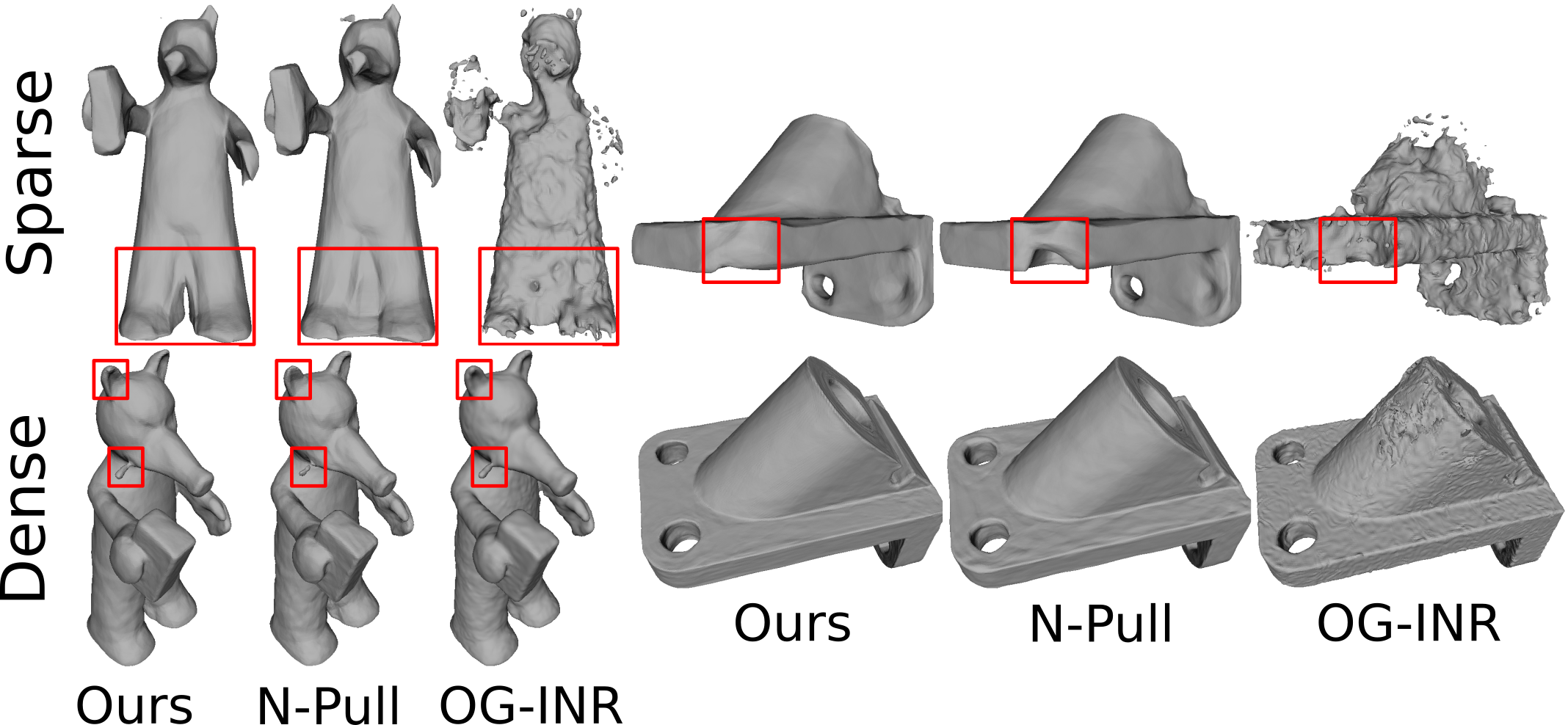}
\caption{SRB \cite{williams2019deep} reconstructions from sparse and dense unoriented inputs.}
\label{fig:srb}
\end{figure}

\begin{table}[h!]
\centering
\scalebox{0.8}{
\begin{tabular}{l|ll|ll}
\hline
& \multicolumn{2}{c|}{Sparse} & \multicolumn{2}{c}{Dense} \\
 & CD1 & HD & CD1 & HD\\ \hline
SPSR  \cite{kazhdan2013screened}   & 2.27 & 21.1 & 1.25  & 22.59 \\
DiGs  \cite{ben2022digs}& 0.68 & 6.05 & \textbf{0.19}  &  \textbf{3.52} \\
OG-INR \cite{koneputugodage2023octree}  & 0.85 & 7.10  &  0.20 & 4.06 \\
NTPS \cite{NeuralTPS}   & 0.73 & 7.78 & - & - \\
N-Pull \cite{ma2020neural} & 0.58 & 8.90 & 0.23 & 4.46  \\
Ours  &\textbf{0.49} & \textbf{6.04} & 0.20 & 3.94
 \\ \hline
\end{tabular}}
\caption{Ablation of point cloud density.}
\label{tab:srb}
\end{table}

\noindent \textbf{Point Cloud Density}. 
We utilize the SRB \cite{williams2019deep} benchmark to evaluate the performance of our method across varying point cloud densities. Table \ref{tab:srb} presents comparative results for  sparse ($1024$)  and dense input point clouds . We include results for the competition from OG-INR in the dense setting. Our method outperforms the competition in the sparse case and performs comparably to the state-of-the-art in the dense case. The improvement over our baseline is substantial for both sparse and dense inputs, as highlighted visually in Figure \ref{fig:srb}, where we showcase reconstructions for both sparse and dense cases. Notably, we achieve better topologies in the sparse case and enhanced, more accurate details in the dense case, as indicated by the red boxes. These results underscore the utility and advantages of our contribution, even in the dense setting. 

\begin{figure}[t!]
    \vspace{-5pt}
    \centering
    \begin{minipage}{0.45\linewidth}
        \centering
        \includegraphics[width=1.0\linewidth]{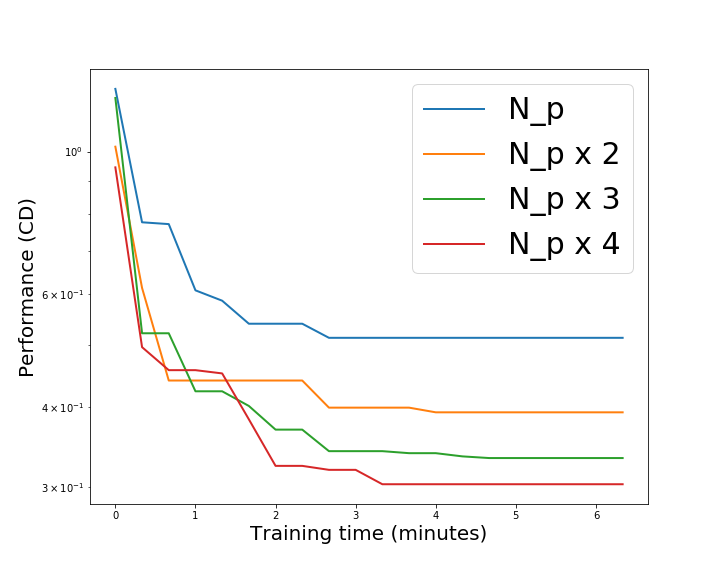}
        \vspace{-25pt}
        \caption{\footnotesize Performance over time (varying input size).}
        \label{fig:input}
    \end{minipage}%
    \hfill
    \begin{minipage}{0.45\linewidth}
        \centering
        \includegraphics[width=1.0\linewidth]{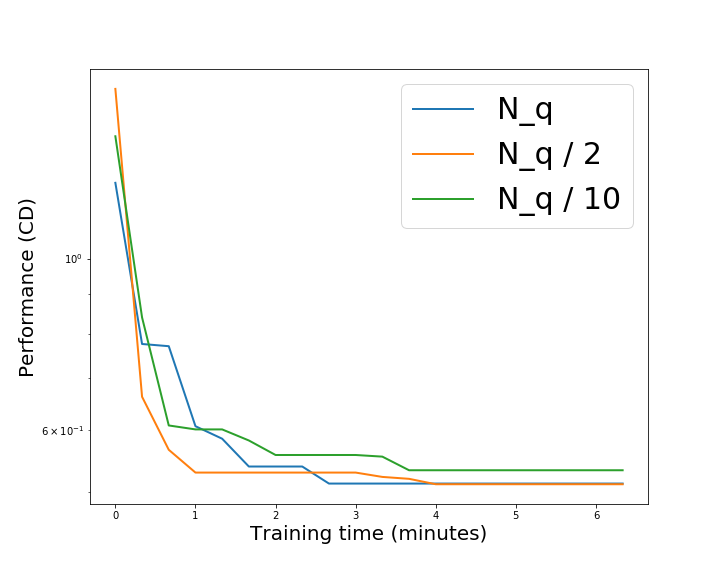}
        \vspace{-25pt}
        \caption{\footnotesize \footnotesize Performance over time (varying query set size).}
        \label{fig:query}
    \end{minipage}
\vspace{-5pt} 
\end{figure}

\noindent\textbf{Running Time and Performance against Number of Queries and Input Points}. Fig.\ref{fig:input} shows our best performance over time (minutes) for various input point cloud sizes. Models using larger inputs reach the baseline performance at earlier times. 
Fig.\ref{fig:query} shows our best performance over time while decreasing the total number of query samples. Performance does not deteriorate drastically under less queries. These Experiments were performed using shape Gargoyle of benchmark SRB.

\noindent \textbf{Robustness to Noise} We evaluate here (Tab.\ref{tab:noise}) our method with various noise levels on class Table of ShapeNet. Performance remains reasonable even under heavily corrupted inputs, which confirms our qualitative superiority to competition in Fig.\ref{fig:sn}, and our resilience to noise.
\begin{table}[t!]
\centering
\scalebox{0.8}{
\begin{tabular}{ cccc }
\hline
 noise std. dev. & 0 & 0.005  & 0.025 \\ 
\hline
 CD1 $\downarrow$ & 0.56 & 0.77 & 2.16 \\  
 NC $\uparrow$ & 0.93 & 0.89 & 0.68 \\ 
\hline
\end{tabular}}
\caption{Noise ablation.}
\label{tab:noise}
\end{table}

\section{Limitations}

We notice that excess regularization can cause reconstructions to be overly smooth
in some cases, which can affect our NC numbers is some benchmarks. This can be improved by tuning the hyperparameters of the method per benchmark or per object/scene.



\section{Conclusion}
\label{sec:conc}

We presented a method for implicit shape reconstruction from sparse, noisy and unoriented point cloud. Our results demonstrate that occupancy fields offer and effective implicit shape representation for this task, and that they can be learned successfully even in the unsupervised setting. Under our carefully designed losses, the occupancy outperforms state-of-the-art methods based on SDF, as-well-as other occupancy based baselines that we show in ablation studies.

{
    \small
    \bibliography{main}

\begin{thebibliography}{98}
\providecommand{\natexlab}[1]{#1}
\providecommand{\url}[1]{\texttt{#1}}
\expandafter\ifx\csname urlstyle\endcsname\relax
  \providecommand{\doi}[1]{doi: #1}\else
  \providecommand{\doi}{doi: \begingroup \urlstyle{rm}\Url}\fi

\bibitem[Aliev et~al.(2020)Aliev, Sevastopolsky, Kolos, Ulyanov, and
  Lempitsky]{aliev2020neural}
Kara-Ali Aliev, Artem Sevastopolsky, Maria Kolos, Dmitry Ulyanov, and Victor
  Lempitsky.
\newblock Neural point-based graphics.
\newblock In \emph{Computer Vision--ECCV 2020: 16th European Conference,
  Glasgow, UK, August 23--28, 2020, Proceedings, Part XXII 16}, pages 696--712.
  Springer, 2020.

\bibitem[Amenta et~al.(2001)Amenta, Choi, and Kolluri]{amenta2001power}
Nina Amenta, Sunghee Choi, and Ravi~Krishna Kolluri.
\newblock The power crust, unions of balls, and the medial axis transform.
\newblock \emph{CG}, 2001.

\bibitem[Atzmon and Lipman(2020{\natexlab{a}})]{atzmon2020sal}
Matan Atzmon and Yaron Lipman.
\newblock Sal: Sign agnostic learning of shapes from raw data.
\newblock In \emph{CVPR}, 2020{\natexlab{a}}.

\bibitem[Atzmon and Lipman(2020{\natexlab{b}})]{atzmon2020sald}
Matan Atzmon and Yaron Lipman.
\newblock Sald: Sign agnostic learning with derivatives.
\newblock In \emph{ICML}, 2020{\natexlab{b}}.

\bibitem[Ben-Israel(1966)]{ben1966newton}
Adi Ben-Israel.
\newblock A newton-raphson method for the solution of systems of equations.
\newblock In \emph{Journal of Mathematical analysis and applications}, 1966.

\bibitem[Ben-Shabat et~al.(2022)Ben-Shabat, Koneputugodage, and
  Gould]{ben2022digs}
Yizhak Ben-Shabat, Chamin~Hewa Koneputugodage, and Stephen Gould.
\newblock Digs: Divergence guided shape implicit neural representation for
  unoriented point clouds.
\newblock In \emph{CVPR}, 2022.

\bibitem[Bernardini et~al.(1999)Bernardini, Mittleman, Rushmeier, Silva, and
  Taubin]{bernardini1999ball}
Fausto Bernardini, Joshua Mittleman, Holly Rushmeier, Claudio Silva, and
  Gabriel Taubin.
\newblock The ball-pivoting algorithm for surface reconstruction.
\newblock \emph{TVCG}, 1999.

\bibitem[Bogo et~al.(2014)Bogo, Romero, Loper, and Black]{Bogo:CVPR:2014}
Federica Bogo, Javier Romero, Matthew Loper, and Michael~J. Black.
\newblock {FAUST}: Dataset and evaluation for {3D} mesh registration.
\newblock In \emph{CVPR}, 2014.

\bibitem[Boulch and Marlet(2022)]{boulch2022poco}
Alexandre Boulch and Renaud Marlet.
\newblock Poco: Point convolution for surface reconstruction.
\newblock In \emph{Proceedings of the IEEE/CVF Conference on Computer Vision
  and Pattern Recognition}, pages 6302--6314, 2022.

\bibitem[Boulch et~al.(2021)Boulch, Langlois, Puy, and
  Marlet]{boulch2021needrop}
Alexandre Boulch, Pierre-Alain Langlois, Gilles Puy, and Renaud Marlet.
\newblock Needrop: Self-supervised shape representation from sparse point
  clouds using needle dropping.
\newblock In \emph{2021 International Conference on 3D Vision (3DV)}, pages
  940--950. IEEE, 2021.

\bibitem[Carr et~al.(2001)Carr, Beatson, Cherrie, Mitchell, Fright, McCallum,
  and Evans]{carr2001reconstruction}
Jonathan~C Carr, Richard~K Beatson, Jon~B Cherrie, Tim~J Mitchell, W~Richard
  Fright, Bruce~C McCallum, and Tim~R Evans.
\newblock Reconstruction and representation of 3d objects with radial basis
  functions.
\newblock In \emph{SIGGRAPH}, 2001.

\bibitem[Cazals and Giesen(2006)]{cazals2006delaunay}
Fr{\'e}d{\'e}ric Cazals and Joachim Giesen.
\newblock \emph{Effective Computational Geometry for Curves and Surfaces}.
\newblock 2006.

\bibitem[Chabra et~al.(2020)Chabra, Lenssen, Ilg, Schmidt, Straub, Lovegrove,
  and Newcombe]{chabra2020deep}
Rohan Chabra, Jan~E Lenssen, Eddy Ilg, Tanner Schmidt, Julian Straub, Steven
  Lovegrove, and Richard Newcombe.
\newblock Deep local shapes: Learning local sdf priors for detailed 3d
  reconstruction.
\newblock In \emph{ECCV}, 2020.

\bibitem[Chan et~al.(2022)Chan, Lin, Chan, Nagano, Pan, De~Mello, Gallo,
  Guibas, Tremblay, Khamis, et~al.]{chan2022efficient}
Eric~R Chan, Connor~Z Lin, Matthew~A Chan, Koki Nagano, Boxiao Pan, Shalini
  De~Mello, Orazio Gallo, Leonidas~J Guibas, Jonathan Tremblay, Sameh Khamis,
  et~al.
\newblock Efficient geometry-aware 3d generative adversarial networks.
\newblock In \emph{Proceedings of the IEEE/CVF Conference on Computer Vision
  and Pattern Recognition}, pages 16123--16133, 2022.

\bibitem[Chang et~al.(2015)Chang, Funkhouser, Guibas, Hanrahan, Huang, Li,
  Savarese, Savva, Song, Su, et~al.]{shapenet}
Angel~X Chang, Thomas Funkhouser, Leonidas Guibas, Pat Hanrahan, Qixing Huang,
  Zimo Li, Silvio Savarese, Manolis Savva, Shuran Song, Hao Su, et~al.
\newblock Shapenet: An information-rich 3d model repository.
\newblock \emph{arXiv preprint arXiv:1512.03012}, 2015.

\bibitem[Chen et~al.(2023{\natexlab{a}})Chen, Han, and Liu]{NeuralTPS}
Chao Chen, Zhizhong Han, and Yu-Shen Liu.
\newblock Unsupervised inference of signed distance functions from single
  sparse point clouds without learning priors.
\newblock In \emph{Proceedings of the IEEE/CVF Conference on Computer Vision
  and Pattern Recognition (CVPR)}, 2023{\natexlab{a}}.

\bibitem[Chen et~al.(2023{\natexlab{b}})Chen, Liu, and Han]{chen2023gridpull}
Chao Chen, Yu-Shen Liu, and Zhizhong Han.
\newblock Gridpull: Towards scalability in learning implicit representations
  from 3d point clouds.
\newblock In \emph{Proceedings of the ieee/cvf international conference on
  computer vision}, pages 18322--18334, 2023{\natexlab{b}}.

\bibitem[Chen and Wang(2022)]{chen2022transinr}
Yinbo Chen and Xiaolong Wang.
\newblock Transformers as meta-learners for implicit neural representations.
\newblock In \emph{European Conference on Computer Vision}, 2022.

\bibitem[Chen and Zhang(2019)]{chen2019learning}
Zhiqin Chen and Hao Zhang.
\newblock Learning implicit fields for generative shape modeling.
\newblock In \emph{CVPR}, 2019.

\bibitem[Chen et~al.(2020)Chen, Tagliasacchi, and Zhang]{chen2020bsp}
Zhiqin Chen, Andrea Tagliasacchi, and Hao Zhang.
\newblock Bsp-net: Generating compact meshes via binary space partitioning.
\newblock In \emph{Proceedings of the IEEE/CVF Conference on Computer Vision
  and Pattern Recognition}, 2020.

\bibitem[Chibane and Pons-Moll(2020)]{chibane2020implicit}
Julian Chibane and Gerard Pons-Moll.
\newblock Implicit feature networks for texture completion from partial 3d
  data.
\newblock In \emph{European Conference on Computer Vision}, pages 717--725.
  Springer, 2020.

\bibitem[Chibane et~al.(2020)Chibane, Mir, and Pons-Moll]{chibane2020neural}
Julian Chibane, Aymen Mir, and Gerard Pons-Moll.
\newblock Neural unsigned distance fields for implicit function learning.
\newblock In \emph{NeurIPS}, 2020.

\bibitem[Deng et~al.(2020)Deng, Genova, Yazdani, Bouaziz, Hinton, and
  Tagliasacchi]{deng2020cvxnet}
Boyang Deng, Kyle Genova, Soroosh Yazdani, Sofien Bouaziz, Geoffrey Hinton, and
  Andrea Tagliasacchi.
\newblock Cvxnet: Learnable convex decomposition.
\newblock In \emph{CVPR}, 2020.

\bibitem[Deprelle et~al.(2019)Deprelle, Groueix, Fisher, Kim, Russell, and
  Aubry]{deprelle2019learning}
Theo Deprelle, Thibault Groueix, Matthew Fisher, Vladimir~G Kim, Bryan~C
  Russell, and Mathieu Aubry.
\newblock Learning elementary structures for 3d shape generation and matching.
\newblock In \emph{NeurIPS}, 2019.

\bibitem[Erler et~al.(2020)Erler, Guerrero, Ohrhallinger, Mitra, and
  Wimmer]{erler2020points2surf}
Philipp Erler, Paul Guerrero, Stefan Ohrhallinger, Niloy~J Mitra, and Michael
  Wimmer.
\newblock Points2surf learning implicit surfaces from point clouds.
\newblock In \emph{ECCV}, 2020.

\bibitem[Fan et~al.(2017)Fan, Su, and Guibas]{fan2017point}
Haoqiang Fan, Hao Su, and Leonidas~J Guibas.
\newblock A point set generation network for 3d object reconstruction from a
  single image.
\newblock In \emph{CVPR}, 2017.

\bibitem[Genova et~al.(2019)Genova, Cole, Vlasic, Sarna, Freeman, and
  Funkhouser]{genova2019learning}
Kyle Genova, Forrester Cole, Daniel Vlasic, Aaron Sarna, William~T Freeman, and
  Thomas Funkhouser.
\newblock Learning shape templates with structured implicit functions.
\newblock In \emph{ICCV}, 2019.

\bibitem[Genova et~al.(2020)Genova, Cole, Sud, Sarna, and
  Funkhouser]{genova2020local}
Kyle Genova, Forrester Cole, Avneesh Sud, Aaron Sarna, and Thomas Funkhouser.
\newblock Local deep implicit functions for 3d shape.
\newblock In \emph{CVPR}, 2020.

\bibitem[Gropp et~al.(2020)Gropp, Yariv, Haim, Atzmon, and
  Lipman]{gropp2020implicit}
Amos Gropp, Lior Yariv, Niv Haim, Matan Atzmon, and Yaron Lipman.
\newblock Implicit geometric regularization for learning shapes.
\newblock In \emph{ICML}, 2020.

\bibitem[Groueix et~al.(2018)Groueix, Fisher, Kim, Russell, and
  Aubry]{groueix2018papier}
Thibault Groueix, Matthew Fisher, Vladimir~G Kim, Bryan~C Russell, and Mathieu
  Aubry.
\newblock A papier-m{\^a}ch{\'e} approach to learning 3d surface generation.
\newblock In \emph{CVPR}, 2018.

\bibitem[Guennebaud and Gross(2007)]{guennebaud2007algebraic}
Ga{\"e}l Guennebaud and Markus Gross.
\newblock Algebraic point set surfaces.
\newblock In \emph{ACM siggraph 2007 papers}, pages 23--es. 2007.

\bibitem[Hart(1996)]{hart1996sphere}
John~C Hart.
\newblock Sphere tracing: A geometric method for the antialiased ray tracing of
  implicit surfaces.
\newblock \emph{The Visual Computer}, 1996.

\bibitem[Huang et~al.(2023)Huang, Gojcic, Atzmon, Litany, Fidler, and
  Williams]{huang2023neural}
Jiahui Huang, Zan Gojcic, Matan Atzmon, Or Litany, Sanja Fidler, and Francis
  Williams.
\newblock Neural kernel surface reconstruction.
\newblock In \emph{Proceedings of the IEEE/CVF Conference on Computer Vision
  and Pattern Recognition}, pages 4369--4379, 2023.

\bibitem[Jain et~al.(2022)Jain, Mildenhall, Barron, Abbeel, and
  Poole]{jain2021dreamfields}
Ajay Jain, Ben Mildenhall, Jonathan~T. Barron, Pieter Abbeel, and Ben Poole.
\newblock Zero-shot text-guided object generation with dream fields.
\newblock 2022.

\bibitem[Jena et~al.(2022)Jena, Multon, and Boukhayma]{jena2022neural}
Shubhendu Jena, Franck Multon, and Adnane Boukhayma.
\newblock Neural mesh-based graphics.
\newblock In \emph{European Conference on Computer Vision}, pages 739--757.
  Springer, 2022.

\bibitem[Jiang et~al.(2020)Jiang, Sud, Makadia, Huang, Nie{\ss}ner, Funkhouser,
  et~al.]{jiang2020local}
Chiyu Jiang, Avneesh Sud, Ameesh Makadia, Jingwei Huang, Matthias Nie{\ss}ner,
  Thomas Funkhouser, et~al.
\newblock Local implicit grid representations for 3d scenes.
\newblock In \emph{CVPR}, 2020.

\bibitem[Kato et~al.(2018)Kato, Ushiku, and Harada]{kato2018neural}
Hiroharu Kato, Yoshitaka Ushiku, and Tatsuya Harada.
\newblock Neural 3d mesh renderer.
\newblock In \emph{CVPR}, 2018.

\bibitem[Kazhdan and Hoppe(2013)]{kazhdan2013screened}
Michael Kazhdan and Hugues Hoppe.
\newblock Screened poisson surface reconstruction.
\newblock \emph{TOG}, 2013.

\bibitem[Kerbl et~al.(2023)Kerbl, Kopanas, Leimk{\"u}hler, and
  Drettakis]{kerbl20233d}
Bernhard Kerbl, Georgios Kopanas, Thomas Leimk{\"u}hler, and George Drettakis.
\newblock 3d gaussian splatting for real-time radiance field rendering.
\newblock \emph{ACM Transactions on Graphics}, 42\penalty0 (4):\penalty0 1--14,
  2023.

\bibitem[Kolluri(2008)]{kolluri2008provably}
Ravikrishna Kolluri.
\newblock Provably good moving least squares.
\newblock \emph{TALG}, 2008.

\bibitem[Koneputugodage et~al.(2023)Koneputugodage, Ben-Shabat, and
  Gould]{koneputugodage2023octree}
Chamin~Hewa Koneputugodage, Yizhak Ben-Shabat, and Stephen Gould.
\newblock Octree guided unoriented surface reconstruction.
\newblock In \emph{Proceedings of the IEEE/CVF Conference on Computer Vision
  and Pattern Recognition}, pages 16717--16726, 2023.

\bibitem[Li et~al.(2023{\natexlab{a}})Li, Multon, and
  Boukhayma]{li2023learning}
Qian Li, Franck Multon, and Adnane Boukhayma.
\newblock Learning generalizable light field networks from few images.
\newblock In \emph{ICASSP 2023-2023 IEEE International Conference on Acoustics,
  Speech and Signal Processing (ICASSP)}, pages 1--5. IEEE, 2023{\natexlab{a}}.

\bibitem[Li et~al.(2023{\natexlab{b}})Li, Multon, and
  Boukhayma]{li2023regularizing}
Qian Li, Franck Multon, and Adnane Boukhayma.
\newblock Regularizing neural radiance fields from sparse rgb-d inputs.
\newblock In \emph{2023 IEEE International Conference on Image Processing
  (ICIP)}, pages 2320--2324. IEEE, 2023{\natexlab{b}}.

\bibitem[Li et~al.(2022)Li, Wen, Liu, Su, and Han]{li2022learning}
Tianyang Li, Xin Wen, Yu-Shen Liu, Hua Su, and Zhizhong Han.
\newblock Learning deep implicit functions for 3d shapes with dynamic code
  clouds.
\newblock In \emph{Proceedings of the IEEE/CVF Conference on Computer Vision
  and Pattern Recognition}, pages 12840--12850, 2022.

\bibitem[Lin et~al.(2022)Lin, Xiao, Shi, and Wang]{lin2022surface}
Siyou Lin, Dong Xiao, Zuoqiang Shi, and Bin Wang.
\newblock Surface reconstruction from point clouds without normals by
  parametrizing the gauss formula.
\newblock \emph{ACM Transactions on Graphics}, 42\penalty0 (2):\penalty0 1--19,
  2022.

\bibitem[Lionar et~al.(2021)Lionar, Emtsev, Svilarkovic, and
  Peng]{lionar2021dynamic}
Stefan Lionar, Daniil Emtsev, Dusan Svilarkovic, and Songyou Peng.
\newblock Dynamic plane convolutional occupancy networks.
\newblock In \emph{Proceedings of the IEEE/CVF Winter Conference on
  Applications of Computer Vision}, pages 1829--1838, 2021.

\bibitem[Lipman(2021)]{lipman2021phase}
Yaron Lipman.
\newblock Phase transitions, distance functions, and implicit neural
  representations.
\newblock In \emph{ICML}, 2021.

\bibitem[Liu et~al.(2018)Liu, Yang, Ceylan, Yumer, and
  Furukawa]{liu2018planenet}
Chen Liu, Jimei Yang, Duygu Ceylan, Ersin Yumer, and Yasutaka Furukawa.
\newblock Planenet: Piece-wise planar reconstruction from a single rgb image.
\newblock In \emph{CVPR}, 2018.

\bibitem[Liu et~al.(2022)Liu, Williams, Jacobson, Fidler, and
  Litany]{liu2022learning}
Hsueh-Ti~Derek Liu, Francis Williams, Alec Jacobson, Sanja Fidler, and Or
  Litany.
\newblock Learning smooth neural functions via lipschitz regularization.
\newblock \emph{arXiv preprint arXiv:2202.08345}, 2022.

\bibitem[Liu et~al.(2020)Liu, Zhang, and Su]{liu2020meshing}
Minghua Liu, Xiaoshuai Zhang, and Hao Su.
\newblock Meshing point clouds with predicted intrinsic-extrinsic ratio
  guidance.
\newblock In \emph{ECCV}, 2020.

\bibitem[Liu et~al.(2021)Liu, Guo, Pan, Wang, Tong, and Liu]{liu2021deep}
Shi-Lin Liu, Hao-Xiang Guo, Hao Pan, Peng-Shuai Wang, Xin Tong, and Yang Liu.
\newblock Deep implicit moving least-squares functions for 3d reconstruction.
\newblock In \emph{CVPR}, 2021.

\bibitem[Lorensen and Cline(1987)]{lorensen1987marching}
William~E Lorensen and Harvey~E Cline.
\newblock Marching cubes: A high resolution 3d surface construction algorithm.
\newblock In \emph{SIGGRAPH}, 1987.

\bibitem[Ma et~al.(2021)Ma, Han, Liu, and Zwicker]{ma2020neural}
Baorui Ma, Zhizhong Han, Yu-Shen Liu, and Matthias Zwicker.
\newblock Neural-pull: Learning signed distance functions from point clouds by
  learning to pull space onto surfaces.
\newblock In \emph{ICML}, 2021.

\bibitem[Ma et~al.(2022{\natexlab{a}})Ma, Liu, and Han]{ma2022reconstructing}
Baorui Ma, Yu-Shen Liu, and Zhizhong Han.
\newblock Reconstructing surfaces for sparse point clouds with on-surface
  priors.
\newblock In \emph{Proceedings of the IEEE/CVF Conference on Computer Vision
  and Pattern Recognition}, pages 6315--6325, 2022{\natexlab{a}}.

\bibitem[Ma et~al.(2022{\natexlab{b}})Ma, Liu, Zwicker, and Han]{ma2022surface}
Baorui Ma, Yu-Shen Liu, Matthias Zwicker, and Zhizhong Han.
\newblock Surface reconstruction from point clouds by learning predictive
  context priors.
\newblock In \emph{Proceedings of the IEEE/CVF Conference on Computer Vision
  and Pattern Recognition}, pages 6326--6337, 2022{\natexlab{b}}.

\bibitem[Mercier et~al.(2022)Mercier, Lescoat, Roussillon, Boubekeur, and
  Thiery]{mercier2022moving}
Corentin Mercier, Thibault Lescoat, Pierre Roussillon, Tamy Boubekeur, and
  Jean-Marc Thiery.
\newblock Moving level-of-detail surfaces.
\newblock \emph{ACM Transactions on Graphics (TOG)}, 41\penalty0 (4):\penalty0
  1--10, 2022.

\bibitem[Mescheder et~al.(2019)Mescheder, Oechsle, Niemeyer, Nowozin, and
  Geiger]{mescheder2019occupancy}
Lars Mescheder, Michael Oechsle, Michael Niemeyer, Sebastian Nowozin, and
  Andreas Geiger.
\newblock Occupancy networks: Learning 3d reconstruction in function space.
\newblock In \emph{Proceedings of the IEEE/CVF conference on computer vision
  and pattern recognition}, pages 4460--4470, 2019.

\bibitem[Mildenhall et~al.(2020)Mildenhall, Srinivasan, Tancik, Barron,
  Ramamoorthi, and Ng]{mildenhall2020nerf}
Ben Mildenhall, Pratul~P Srinivasan, Matthew Tancik, Jonathan~T Barron, Ravi
  Ramamoorthi, and Ren Ng.
\newblock Nerf: Representing scenes as neural radiance fields for view
  synthesis.
\newblock In \emph{ECCV}, 2020.

\bibitem[Ouasfi and Boukhayma(2022)]{ouasfi2022few}
Amine Ouasfi and Adnane Boukhayma.
\newblock Few'zero level set'-shot learning of shape signed distance functions
  in feature space.
\newblock In \emph{ECCV}, 2022.

\bibitem[Ouasfi and Boukhayma(2023)]{ouasfi2023Robustifying}
Amine Ouasfi and Adnane Boukhayma.
\newblock Robustifying generalizable implicit shape networks with a tunable
  non-parametric model.
\newblock In \emph{NeurIPS}, 2023.

\bibitem[Ouasfi and Boukhayma(2024)]{ouasfi2024Mixing}
Amine Ouasfi and Adnane Boukhayma.
\newblock Mixing-denoising generalizable occupancy networks.
\newblock In \emph{3DV}, 2024.

\bibitem[Palmer et~al.(2022)Palmer, Smirnov, Wang, Chern, and
  Solomon]{palmer2022deepcurrents}
David Palmer, Dmitriy Smirnov, Stephanie Wang, Albert Chern, and Justin
  Solomon.
\newblock Deepcurrents: Learning implicit representations of shapes with
  boundaries.
\newblock In \emph{Proceedings of the IEEE/CVF Conference on Computer Vision
  and Pattern Recognition}, pages 18665--18675, 2022.

\bibitem[Park et~al.(2019)Park, Florence, Straub, Newcombe, and
  Lovegrove]{park2019deepsdf}
Jeong~Joon Park, Peter Florence, Julian Straub, Richard Newcombe, and Steven
  Lovegrove.
\newblock Deepsdf: Learning continuous signed distance functions for shape
  representation.
\newblock In \emph{CVPR}, 2019.

\bibitem[Paszke et~al.(2019)Paszke, Gross, Massa, Lerer, Bradbury, Chanan,
  Killeen, Lin, Gimelshein, Antiga, et~al.]{paszke2019pytorch}
Adam Paszke, Sam Gross, Francisco Massa, Adam Lerer, James Bradbury, Gregory
  Chanan, Trevor Killeen, Zeming Lin, Natalia Gimelshein, Luca Antiga, et~al.
\newblock Pytorch: An imperative style, high-performance deep learning library.
\newblock \emph{NeurIPS}, 2019.

\bibitem[Peng et~al.(2020)Peng, Niemeyer, Mescheder, Pollefeys, and
  Geiger]{peng2020convolutional}
Songyou Peng, Michael Niemeyer, Lars Mescheder, Marc Pollefeys, and Andreas
  Geiger.
\newblock Convolutional occupancy networks.
\newblock In \emph{European Conference on Computer Vision}, pages 523--540.
  Springer, 2020.

\bibitem[Peng et~al.(2021)Peng, Jiang, Liao, Niemeyer, Pollefeys, and
  Geiger]{peng2021shape}
Songyou Peng, Chiyu Jiang, Yiyi Liao, Michael Niemeyer, Marc Pollefeys, and
  Andreas Geiger.
\newblock Shape as points: A differentiable poisson solver.
\newblock \emph{Advances in Neural Information Processing Systems},
  34:\penalty0 13032--13044, 2021.

\bibitem[Qi et~al.(2017)Qi, Su, Mo, and Guibas]{qi2017pointnet}
Charles~R Qi, Hao Su, Kaichun Mo, and Leonidas~J Guibas.
\newblock Pointnet: Deep learning on point sets for 3d classification and
  segmentation.
\newblock In \emph{CVPR}, 2017.

\bibitem[Rakotosaona et~al.(2021)Rakotosaona, Aigerman, Mitra, Ovsjanikov, and
  Guerrero]{rakotosaona2021differentiable}
Marie-Julie Rakotosaona, Noam Aigerman, Niloy Mitra, Maks Ovsjanikov, and Paul
  Guerrero.
\newblock Differentiable surface triangulation.
\newblock In \emph{SIGGRAPH Asia}, 2021.

\bibitem[Riegler et~al.(2017)Riegler, Osman~Ulusoy, and
  Geiger]{riegler2017octnet}
Gernot Riegler, Ali Osman~Ulusoy, and Andreas Geiger.
\newblock Octnet: Learning deep 3d representations at high resolutions.
\newblock In \emph{CVPR}, 2017.

\bibitem[Sch{\"o}lkopf et~al.(2004)Sch{\"o}lkopf, Giesen, and
  Spalinger]{scholkopf2004kernel}
Bernhard Sch{\"o}lkopf, Joachim Giesen, and Simon Spalinger.
\newblock Kernel methods for implicit surface modeling.
\newblock In \emph{NeurIPS}, 2004.

\bibitem[Sch\"{o}nberger and Frahm(2016)]{schoenberger2016sfm}
Johannes~Lutz Sch\"{o}nberger and Jan-Michael Frahm.
\newblock Structure-from-motion revisited.
\newblock In \emph{Conference on Computer Vision and Pattern Recognition
  (CVPR)}, 2016.

\bibitem[Sch\"{o}nberger et~al.(2016)Sch\"{o}nberger, Zheng, Pollefeys, and
  Frahm]{schoenberger2016mvs}
Johannes~Lutz Sch\"{o}nberger, Enliang Zheng, Marc Pollefeys, and Jan-Michael
  Frahm.
\newblock Pixelwise view selection for unstructured multi-view stereo.
\newblock In \emph{European Conference on Computer Vision (ECCV)}, 2016.

\bibitem[Settles(2009)]{settles2009active}
Burr Settles.
\newblock Active learning literature survey.
\newblock 2009.

\bibitem[Shannon(1948)]{shannon1948mathematical}
Claude~Elwood Shannon.
\newblock A mathematical theory of communication.
\newblock \emph{The Bell system technical journal}, 27\penalty0 (3):\penalty0
  379--423, 1948.

\bibitem[Sitzmann et~al.(2019)Sitzmann, Zollhoefer, and
  Wetzstein]{NEURIPS2019_b5dc4e5d}
Vincent Sitzmann, Michael Zollhoefer, and Gordon Wetzstein.
\newblock Scene representation networks: Continuous 3d-structure-aware neural
  scene representations.
\newblock In \emph{NeurIPS}, 2019.

\bibitem[Sitzmann et~al.(2020{\natexlab{a}})Sitzmann, Chan, Tucker, Snavely,
  and Wetzstein]{sitzmann2020metasdf}
Vincent Sitzmann, Eric~R Chan, Richard Tucker, Noah Snavely, and Gordon
  Wetzstein.
\newblock Metasdf: Meta-learning signed distance functions.
\newblock In \emph{NeurIPS}, 2020{\natexlab{a}}.

\bibitem[Sitzmann et~al.(2020{\natexlab{b}})Sitzmann, Martel, Bergman, Lindell,
  and Wetzstein]{sitzmann2020implicit}
Vincent Sitzmann, Julien Martel, Alexander Bergman, David Lindell, and Gordon
  Wetzstein.
\newblock Implicit neural representations with periodic activation functions.
\newblock In \emph{NeurIPS}, 2020{\natexlab{b}}.

\bibitem[Sitzmann et~al.(2021)Sitzmann, Rezchikov, Freeman, Tenenbaum, and
  Durand]{sitzmann2021light}
Vincent Sitzmann, Semon Rezchikov, William~T Freeman, Joshua~B Tenenbaum, and
  Fredo Durand.
\newblock Light field networks: Neural scene representations with
  single-evaluation rendering.
\newblock In \emph{NeurIPS}, 2021.

\bibitem[Takikawa et~al.(2021)Takikawa, Litalien, Yin, Kreis, Loop,
  Nowrouzezahrai, Jacobson, McGuire, and Fidler]{takikawa2021neural}
Towaki Takikawa, Joey Litalien, Kangxue Yin, Karsten Kreis, Charles Loop, Derek
  Nowrouzezahrai, Alec Jacobson, Morgan McGuire, and Sanja Fidler.
\newblock Neural geometric level of detail: Real-time rendering with implicit
  3d shapes.
\newblock In \emph{CVPR}, 2021.

\bibitem[Tang et~al.(2021)Tang, Lei, Xu, Ma, Jia, and Zhang]{tang2021sa}
Jiapeng Tang, Jiabao Lei, Dan Xu, Feiying Ma, Kui Jia, and Lei Zhang.
\newblock Sa-convonet: Sign-agnostic optimization of convolutional occupancy
  networks.
\newblock In \emph{Proceedings of the IEEE/CVF International Conference on
  Computer Vision}, pages 6504--6513, 2021.

\bibitem[Tatarchenko et~al.(2017)Tatarchenko, Dosovitskiy, and
  Brox]{tatarchenko2017octree}
Maxim Tatarchenko, Alexey Dosovitskiy, and Thomas Brox.
\newblock Octree generating networks: Efficient convolutional architectures for
  high-resolution 3d outputs.
\newblock In \emph{ICCV}, 2017.

\bibitem[Tretschk et~al.(2020)Tretschk, Tewari, Golyanik, Zollh{\"o}fer, Stoll,
  and Theobalt]{tretschk2020patchnets}
Edgar Tretschk, Ayush Tewari, Vladislav Golyanik, Michael Zollh{\"o}fer,
  Carsten Stoll, and Christian Theobalt.
\newblock Patchnets: Patch-based generalizable deep implicit 3d shape
  representations.
\newblock In \emph{ECCV}, 2020.

\bibitem[Tulsiani et~al.(2017)Tulsiani, Su, Guibas, Efros, and
  Malik]{abstractionTulsiani17}
Shubham Tulsiani, Hao Su, Leonidas~J. Guibas, Alexei~A. Efros, and Jitendra
  Malik.
\newblock Learning shape abstractions by assembling volumetric primitives.
\newblock In \emph{CVPR}, 2017.

\bibitem[Wang et~al.(2018)Wang, Zhang, Li, Fu, Liu, and
  Jiang]{wang2018pixel2mesh}
Nanyang Wang, Yinda Zhang, Zhuwen Li, Yanwei Fu, Wei Liu, and Yu-Gang Jiang.
\newblock Pixel2mesh: Generating 3d mesh models from single rgb images.
\newblock In \emph{ECCV}, 2018.

\bibitem[Wang et~al.(2021{\natexlab{a}})Wang, Liu, Liu, Theobalt, Komura, and
  Wang]{wang2021neus}
Peng Wang, Lingjie Liu, Yuan Liu, Christian Theobalt, Taku Komura, and Wenping
  Wang.
\newblock Neus: Learning neural implicit surfaces by volume rendering for
  multi-view reconstruction.
\newblock \emph{arXiv preprint arXiv:2106.10689}, 2021{\natexlab{a}}.

\bibitem[Wang et~al.(2017)Wang, Liu, Guo, Sun, and Tong]{wang2017cnn}
Peng-Shuai Wang, Yang Liu, Yu-Xiao Guo, Chun-Yu Sun, and Xin Tong.
\newblock O-cnn: Octree-based convolutional neural networks for 3d shape
  analysis.
\newblock \emph{TOG}, 2017.

\bibitem[Wang et~al.(2021{\natexlab{b}})Wang, Mihajlovic, Ma, Geiger, and
  Tang]{wang2021metaavatar}
Shaofei Wang, Marko Mihajlovic, Qianli Ma, Andreas Geiger, and Siyu Tang.
\newblock Metaavatar: Learning animatable clothed human models from few depth
  images.
\newblock \emph{Advances in Neural Information Processing Systems},
  34:\penalty0 2810--2822, 2021{\natexlab{b}}.

\bibitem[Williams et~al.(2019)Williams, Schneider, Silva, Zorin, Bruna, and
  Panozzo]{williams2019deep}
Francis Williams, Teseo Schneider, Claudio Silva, Denis Zorin, Joan Bruna, and
  Daniele Panozzo.
\newblock Deep geometric prior for surface reconstruction.
\newblock In \emph{CVPR}, 2019.

\bibitem[Williams et~al.(2021)Williams, Trager, Bruna, and
  Zorin]{williams2021neural}
Francis Williams, Matthew Trager, Joan Bruna, and Denis Zorin.
\newblock Neural splines: Fitting 3d surfaces with infinitely-wide neural
  networks.
\newblock In \emph{CVPR}, 2021.

\bibitem[Williams et~al.(2022)Williams, Gojcic, Khamis, Zorin, Bruna, Fidler,
  and Litany]{williams2022neural}
Francis Williams, Zan Gojcic, Sameh Khamis, Denis Zorin, Joan Bruna, Sanja
  Fidler, and Or Litany.
\newblock Neural fields as learnable kernels for 3d reconstruction.
\newblock In \emph{CVPR}, 2022.

\bibitem[Wu et~al.(2016)Wu, Zhang, Xue, Freeman, and Tenenbaum]{wu2016learning}
Jiajun Wu, Chengkai Zhang, Tianfan Xue, William~T Freeman, and Joshua~B
  Tenenbaum.
\newblock Learning a probabilistic latent space of object shapes via 3d
  generative-adversarial modeling.
\newblock In \emph{NeurIPS}, 2016.

\bibitem[Wu et~al.(2015)Wu, Song, Khosla, Yu, Zhang, Tang, and Xiao]{wu20153d}
Zhirong Wu, Shuran Song, Aditya Khosla, Fisher Yu, Linguang Zhang, Xiaoou Tang,
  and Jianxiong Xiao.
\newblock 3d shapenets: A deep representation for volumetric shapes.
\newblock In \emph{CVPR}, 2015.

\bibitem[Yariv et~al.(2021)Yariv, Gu, Kasten, and Lipman]{yariv2021volume}
Lior Yariv, Jiatao Gu, Yoni Kasten, and Yaron Lipman.
\newblock Volume rendering of neural implicit surfaces.
\newblock \emph{Advances in Neural Information Processing Systems},
  34:\penalty0 4805--4815, 2021.

\bibitem[Yavartanoo et~al.(2021)Yavartanoo, Chung, Neshatavar, and
  Lee]{yavartanoo20213dias}
Mohsen Yavartanoo, Jaeyoung Chung, Reyhaneh Neshatavar, and Kyoung~Mu Lee.
\newblock 3dias: 3d shape reconstruction with implicit algebraic surfaces.
\newblock In \emph{ICCV}, 2021.

\bibitem[Ypma(1995)]{ypma1995historical}
Tjalling~J Ypma.
\newblock Historical development of the newton--raphson method.
\newblock \emph{SIAM review}, 37\penalty0 (4):\penalty0 531--551, 1995.

\bibitem[Zhou et~al.(2022)Zhou, Ma, Yu-Shen, Yi, and Zhizhong]{Zhou2022CAP-UDF}
Junsheng Zhou, Baorui Ma, Liu Yu-Shen, Fang Yi, and Han Zhizhong.
\newblock Learning consistency-aware unsigned distance functions progressively
  from raw point clouds.
\newblock In \emph{Advances in Neural Information Processing Systems
  (NeurIPS)}, 2022.

\bibitem[Zhou and Koltun(2013)]{zhou2013dense}
Qian-Yi Zhou and Vladlen Koltun.
\newblock Dense scene reconstruction with points of interest.
\newblock \emph{ACM Transactions on Graphics (ToG)}, 32\penalty0 (4):\penalty0
  1--8, 2013.

\bibitem[Zou et~al.(2017)Zou, Yumer, Yang, Ceylan, and Hoiem]{zou20173d}
Chuhang Zou, Ersin Yumer, Jimei Yang, Duygu Ceylan, and Derek Hoiem.
\newblock 3d-prnn: Generating shape primitives with recurrent neural networks.
\newblock In \emph{CVPR}, 2017.

\end{thebibliography}
    \bibliographystyle{ieeenat_fullname}
}

\clearpage
\setcounter{page}{1}
\maketitlesupplementary
\appendix

\section{Additional Implementation Details}

Unless stated differently, We employ publicly accessible official implementations of established methods in our work. Specifically, for SPSR, in accordance with our peer-reviewed competition practices, we utilize publicly available implementations provided by Open3D and Pymeshlab. We select the superior result between the two and tune its hyperparameters including grid searches for parameters such as octree depth and the number of nearest neighbors utilized in constructing the Riemannian graph for normal orientation propagation. It is important to highlight that these libraries feature a normal estimation algorithm grounded in local point cloud co-variance estimation, coupled with normal orientation propagation employing minimum spanning trees.

\section{Metrics}

Following the definitions from \cite{boulch2022poco} and \cite{williams2019deep}, we present here the formal definitions for the metrics that we use for evaluation in the main submission. We denote by $\mathcal{S}$ and  $\hat{\mathcal{S}}$ the ground truth and predicted mesh respectively. We follow \cite{NeuralTPS} to approximate all metrics  with 100k samples from $\mathcal{S}$ and $\hat{\mathcal{S}}$ for ShapeNet and Faust and with 1M samples for 3Dscene. For SRB, we use 1M samples  following \cite{ben2022digs} and \cite{koneputugodage2023octree}. 

\paragraph{Chamfer Distance (CD$_1$)} The L$_1$ Chamfer distance is based on the two-ways nearest neighbor distance: 
$$\mathrm{CD}_1=\frac{1}{2|\mathcal{S}|} \sum_{v \in \mathcal{S}} \min _{\hat{v} \in \hat{\mathcal{S}}}\|v-\hat{v}\|_2+\frac{1}{2|\hat{\mathcal{S}}|} \sum_{\hat{v} \in \hat{\mathcal{S}}} \min _{v \in \mathcal{S}}\|\hat{v}-v\|_2.$$

\paragraph{Chamfer Distance (CD$_2$)} The L$_2$ Chamfer distance is based on the two-ways nearest neighbor squared distance: 
$$\mathrm{CD}_2=\frac{1}{2|\mathcal{S}|} \sum_{v \in \mathcal{S}} \min _{\hat{v} \in \hat{\mathcal{S}}}\|v-\hat{v}\|_2^2+\frac{1}{2|\hat{\mathcal{S}}|} \sum_{\hat{v} \in \hat{\mathcal{S}}} \min _{v \in \mathcal{S}}\|\hat{v}-v\|_2^2.$$

\paragraph{F-Score (FS)} For a given threshold $\tau$, the F-score between  the meshes $\mathcal{S}$ and $\hat{\mathcal{S}}$ is defined as:
$$
\mathrm{FS}\left(\tau, \mathcal{S}, \hat{\mathcal{S}}\right)=\frac{2 \text { Recall} \cdot \text{Precision }}{\text { Recall }+\text { Precision }},
$$

where
$$
\begin{array}{r}
\operatorname{Recall}\left(\tau, \mathcal{S}, \hat{\mathcal{S}}\right)=\mid\left\{v \in \mathcal{S} \text {, s.t. } \min _{\hat{v} \in \hat{ \mathcal{S} }} \left\|v-\hat{v}\|_2\right<\tau\right\} \mid ,\\
\operatorname{Precision}\left(\tau, \mathcal{S}, \hat{\mathcal{S}}\right)=\mid\left\{\hat{v} \in \hat{\mathcal{S} }\text {, s.t. } \min _{v \in  \mathcal{S} } \left\|v-\hat{v}\|_2\right<\tau\right\} \mid .\\
\end{array}
$$
Following \cite{mescheder2019occupancy} and \cite{peng2020convolutional}, we set $\tau$ to $0.01$.

\paragraph{Normal consistency (NC)} We denote here by $n_v$ the normal at a point $v$ in $\mathcal{S}$. The normal consistency between two meshes $\mathcal{S}$ and $\hat{\mathcal{S}}$ is defined as: 

$$\mathrm{NC}=\frac{1}{2|\mathcal{S}|} \sum_{v \in \mathcal{S}} n_{v} \cdot n_{\operatorname{closest}(v,\hat{ \mathcal{S}})}+\frac{1}{2|\hat{\mathcal{S}}|} \sum_{\hat{v} \in \hat{\mathcal{S}}} n_{\hat{v}} \cdot n_{\operatorname{closest}(\hat{v}, \mathcal{S})},$$

where  
$$
\operatorname{closest}(v, \hat{\mathcal{S}}) = \operatorname{argmin} _{\hat{v} \in \hat{\mathcal{S}}}\|v-\hat{v}\|_2.
$$

\paragraph{Hausdorff distance (HD)} This metric is defined as follows:

$$d_H = \max\left(\max _{v \in \mathcal{S}}\min _{\hat{v} \in \hat{\mathcal{S}}}\|v-\hat{v}\|_2,\max _{\hat{v} \in \hat{\mathcal{S}}}\min _{v \in \mathcal{S}}\|v-\hat{v}\|_2\right)$$

\end{document}